\documentclass[acmlarge,screen]{acmart}

\AtBeginDocument{%
  \providecommand\BibTeX{{%
    \normalfont B\kern-0.5em{\scshape i\kern-0.25em b}\kern-0.8em\TeX}}}

\setcopyright{rightsretained}
\acmJournal{IMWUT}
\acmYear{2023} \acmVolume{7} \acmNumber{3} \acmArticle{126} \acmMonth{9} \acmPrice{}\acmDOI{10.1145/3610887}

\usepackage{enumitem}
\usepackage{makecell}
\usepackage{multirow}
\usepackage[toc,page]{appendix}
\begin{document}

\title{Privacy against Real-Time Speech Emotion Detection via Acoustic Adversarial Evasion of Machine Learning}


\author{Brian Testa}
\email{bptesta@syr.edu}
\orcid{0000-0003-2349-9564}
\author{Yi Xiao}
\email{yixiao54@syr.edu}
\orcid{0000-0002-5261-5440}
\author{Harshit Sharma}
\email{hsharm04@syr.edu}
\orcid{0000-0002-7016-6220}
\author{Avery Gump}
\email{atgump@syr.edu}
\orcid{0009-0007-9535-3974}
\author{Asif Salekin}
\email{asalekin@syr.edu}
\orcid{0000-0002-0807-8967}
\thanks{Asif Salekin is the corresponding author.}
\affiliation{%
  \institution{Syracuse University}
  \streetaddress{900 South Crouse Ave}
  \city{Syracuse}
  \state{New York}
  \country{USA}
  \postcode{13244}
}

\renewcommand{\shortauthors}{Testa et al.}
\renewcommand{\shorttitle}{Privacy against Real-Time Speech Emotion Detection via Acoustic Adversarial Evasion of Machine Learning}
\begin{abstract}

Smart speaker voice assistants (VAs) such as Amazon Echo and Google Home have been widely adopted due to their seamless integration with smart home devices and the Internet of Things (IoT) technologies. These VA services raise privacy concerns, especially due to their access to our speech. This work considers one such use case: the unaccountable and unauthorized surveillance of a user's emotion via speech emotion recognition (SER). This paper presents DARE-GP, a solution that creates additive noise to mask users' emotional information while preserving the transcription-relevant portions of their speech. DARE-GP does this by using a constrained genetic programming approach to learn the spectral frequency traits that depict target users' emotional content, and then generating a universal adversarial audio perturbation that provides this privacy protection. Unlike existing works, DARE-GP provides: a) real-time protection of previously unheard utterances, b) against previously unseen black-box SER classifiers, c) while protecting speech transcription, and d) does so in a realistic, acoustic environment. Further, this evasion is robust against defenses employed by a knowledgeable adversary. The evaluations in this work culminate with acoustic evaluations against two off-the-shelf commercial smart speakers using a small-form-factor (raspberry pi) integrated with a wake-word system to evaluate the efficacy of its real-world, real-time deployment.

\end{abstract}


  
\maketitle

\section{Introduction}\label{section:introduction}

\textbf{Motivation:} Smart speaker voice assistants (VAs) such as Amazon Echo and Google Home have been widely adopted due to their seamless integration with smart home devices and the Internet of Things (IoT) technologies \cite{varef, terzopoulos2020voice, bentley2018understanding}. However, VA services raise privacy concerns, especially due to their access to our speech, which big tech companies may leverage for monetization, a practice called \textit{surveillance capitalism} \cite{barney_wigmore_2022, laidler_2019, zuboff2015big}. In the context of smart speaker VA use, users have no right or control over their speech recordings once the command speech is uploaded to the cloud \cite{Klobuchar2021, alexaalways}. One possible use for this speech data is surveillance of users' emotional data.

Large tech companies have demonstrated interest in capturing affect information from speech through both explicit service offerings (e.g. - Amazon Halo \cite{Klobuchar2021}), and through research and registered patents for technologies developed for affect assessment from speech acoustic, including Amazon \cite{jin2018voice, fussell_2019},  Microsoft \cite{schwartz_2021}, Apple \cite{kowtha2020detecting}, and Spotify \cite{walker_2022}. This investment in speech emotion recognition (SER) indicates two things: 1) these companies have a motivation for collecting users' emotion data, and 2) these companies cannot simply inspect the transcripts of our smart speaker VA interactions to get this information. This second point is supported by a study published by Mozilla and Yahoo \cite{ammari2019music}, the primary uses of the Google Home and Amazon Alexa, i.e., smart speaker VAs are task commands, and conversational interactions are relatively rare (< 10\%). This point is further supported by a short analysis of the emotional content in the transcripts of common smart speaker commands. The results of this analysis are provided in Appendix \ref{appendix:sentiment}. 

As for companies' motivation to collect this information, the work of Lerner et al. \cite{lerner2015emotion} and many others \cite{frijda1988laws,keltner2014understanding,lazarus1991emotion}, emotion has a profound impact upon decision making. Authors at \emph{The Atlantic} noted these trends in speech emotion recognition patents amongst technology companies and came to a reasonable conclusion: these technologies could be applied for targeted advertising based upon a user’s emotions \cite{fussell_2019}. Such exploitation of users' emotional information is not innocuous. Studies have shown that consumers’ affective states and stress positively affect their impulsive \cite{rodrigues2021factors} and compulsive \cite{zheng2020perceived, roberts2012stress} buying behaviors. Notably, increasing impulsive buying may cause compulsive buying disorder \cite{darrat2016impulse,filomensky2021compulsive}, which may result in substantial debts, legal problems, and personal distress \cite{kumar2021impulse, weinstein2016compulsive}. Furthermore, recent studies have shown \cite{aslam2021impact} that personalized advertisement positively affects impulse buying; hence it can be assumed that personalized advertisements based on users’ affective states or stress would exploit the vulnerability of individuals suffering from such disorders and may even contribute to causing them.

Unauthorized, unaccountable exposure of emotion information is a real concern beyond just commercial use cases. In government, affective computing is beginning to influence law enforcement activities \cite{facerobot, Macaulay2020, nyp2021}. In recent years, United States law enforcement has dramatically increased subpoenas for user interaction recordings from smart speaker companies to assess individuals’ mental states \cite{legalexaminer2020, polce-want-audio}. These law enforcement use cases are especially troubling in the context of significant bias and fairness concerns in affective state recognition \cite{race-bias-emo, manresa2022facial, schmitz2022bias, xu2020investigating}, which may create disparity against the minority populations. 

\par

Many individuals are concerned by the exploitation of their private data by companies or the government. In the US \cite{auxier2019americans} and the UK \cite{Macaulay2020, mcstay2020emotional} many have significant reservations about sharing smart speaker VA audio with external stakeholders such as advertisers and law-enforcement agencies. In the UK, about 52\% of the population has chosen not to partake in VA services due to privacy concerns \cite{mcstay2020emotional}. In the United States, concerns regarding the potential abuses of affective computing \cite{Engler2021, Klobuchar2021}, have escalated as high as the United States Senate. On the world stage, the United Nations Human Rights Council has weighed in \cite{mcstay2020emotional}, acknowledging \textit{"...emotion recognition technologies as an emergent priority in the global right to privacy in the digital age"}.

\par

The purpose of this work is to empower smart speaker VA users to protect their private emotion information. Put another way, this work addresses the question: \textit{Can users exploit the utility of smart speaker VA services while limiting the inadvertent disclosure of emotional information depicted through speakers’ acoustic attributes?} Any answer to this question must address all of the following research questions:

\begin{enumerate}[leftmargin=0.6cm]

    \item \textit{Given a set of users, can an approach deceive a (i) previously unseen black-box SER classifier (ii) without compromising the speech-to-text transcription on (iii) previously unheard utterances (i.e., commands)?}

    \item \textit{How does the performance of this approach compare with state-of-the-art (SotA) audio evasion techniques?}
    
    \item \textit{Can a knowledgeable SER operator defend against this technique?}
    
    \item \textit{Can such protection be granted in an acoustic, real-world scenario with: closed, off-the-shelf (OTS) smart speakers, variable user location related to the smart speaker, and space weight and power (SWAP) \cite{spitzer2017rtca} constraints?}
    \par

\end{enumerate}

A `closed' smart speaker is one that does not provide an API for audio processing before the audio is uploaded to the cloud (discussed in Section \ref{section:threats}). This is a common characteristic across many commercial smart speakers, including Amazon Echo \cite{wallis_2022,Lippett_2023}.

\textbf{Approach:} This paper presents an approach called \textbf{Defeating Acoustic Recognition of Emotion via Genetic Programming (DARE-GP)}. Given a set of users, DARE-GP uses \textit{constrained genetic programming (GP)} to generate a universal adversarial audio perturbation (UAAP) \cite{abdoli2019universal}. For the respective users' speech (even for previously unheard VA commands, i.e., not used during UAAP generation), that single UAAP causes misclassification on previously-unseen black-box SER classifiers while maintaining audio transcription utility (i.e., the constraint); these universal spectral perturbations are called \textbf{Emotion Obfuscating Noises (EONs)}. 
DARE-GP approach is motivated by the insights of Weninger et al. \cite{weninger2013acoustics} and B{\u{a}}lan et al. \cite{bualan2019emotion}, establishing the presence of the \emph{spectral-emotional content relationship} in speech and SER classifiers’ utilization of such relationship in emotion detection.  Furthermore, studies \cite{aloufi2019emotion, aloufi2020paralinguistic, aloufi2020privacy} have shown that the paralinguistic emotional content can be disentangled from the linguistic transcription content in speech, and that SER inference from spectral information of emotionally neutral utterances is quite feasible \cite{makiuchi2021multimodal, baevski2020wav2vec, dupuis2010toronto, hsu2021hubert, livingstone2018ryerson, jackson2014surrey}. Leveraging constrained GP for a target set of users, DARE-GP identifies the spectral attributes through which they most commonly depict their speech emotions and are not related to the depiction of linguistic transcription content. Finally, it generates a single EON that masks such spectral attributes. Hence, while playing simultaneously with users’ speech (even the previously unheard ones), the respective EON masks speakers’ emotional content while preserving its transcription utility.

An effective, real-world emotion protection mechanism should demonstrate all of the following attributes:

\begin{itemize}[leftmargin=0.4cm]

\item \textbf{Audio Utility Preservation} - If a solution disrupts the transcription of audio, then it is not a usable solution; it would make the smart speaker effectively useless. DARE-GP’s EON development explicitly conserves the smart speaker’s ability to transcribe user audio by imposing \textit{constraints} on genetic programming.

\item \textbf{Transferable to Previously Unseen Black-Box Classifiers} - As described in \textbf{Section \ref{section:threats}}, DARE-GP does not have query access \cite{demontis2019adversarial} to the target SER classifier, making it impossible to directly develop a surrogate SER classifier, mimicking the target one. DARE-GP generates noise in the \textit{spectral domain}, which intrinsically makes an EON transferable to \textit{previously unseen SER classifiers} without modification.
 
\item \textbf{Supports Closed, OTS Smart Speaker VA Interaction} - The closed nature of smart speaker processing \cite{Lippett_2023} precludes changes to hardware-implemented features like beamforming or software-based audio processing before the audio is provided to cloud services for processing \cite{wallis_2022}; this limits the scope of possible approaches to deceive a back-end SER classifier.  An EON is played as \textit{additive noise} simultaneously with user speech; this supports real-world integration without requiring any specific smart speaker/VA-specific APIs, interfaces, etc. 

\item \textbf{Real-Time for Previously Unheard Utterances} - Smart speakers' appeal is their ease of use; introducing latency to or requiring replay of user commands would significantly degrade a user’s smart speaker experience. As discussed above, DARE-GP precomputes \textit{a single EON}, which is \textit{universal in nature for a set of users}. Meaning, this additive noise (i.e., EON) can be played in real-time to cause misclassification of previously unheard utterances without any new, on the fly, utterance-specific processing (similar to \cite{li2020advpulse}).

\item \textbf{Effective in an Acoustic Environment} - A solution cannot assume feature-level access to the SER model; as described in Section \ref{section:threats}, interaction with the smart speaker is limited to existing, user-facing interfaces. EONs are developed explicitly for acoustic environments, and do not suffer the issues encountered when attempting to manifest feature-space-developed evasive samples in an acoustic setting \cite{schonherr2020imperio, li2020practical} (see Section \ref{section:evalcompare}).

\item \textbf{Robust versus Knowledgeable Defender} - A solution should assume that the SER operator will have knowledge of its implementation and will attempt to perform accurate emotion inference despite the solution; details about such a defender’s expected knowledge and resources are provided in Section \ref{section:threats}. EONs have been demonstrated to be robust versus multiple possible defenses (see Section \ref{section:defenses}).

\item \textbf{Noninvasive} - Any solution should not be disruptive to a user’s daily life. The impact of an EON to a user is discussed in \textbf{Section \ref{real-alexa-evaluation}}. In addition, in a recent user survey 89.5\% of respondents indicated that the additive noise generated by DARE-GP was either \textit{Noninvasive} or \textit{Completely Inaudible}. Additional details of this survey are provided in Section \ref{section:discussion} and Appendix \ref{appendix:survey}.

\end{itemize}

\begin{figure}[]
\centering
\includegraphics[width=0.5\linewidth]{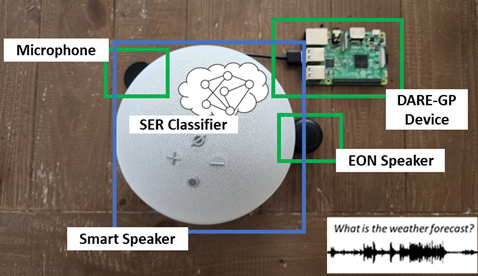}
\caption{DARE-GP employment scenario. An adjacent DARE-GP device can be used to prevent reliable inference by a smart speaker's back-end SER classifier. Green box regions are part of the DARE-GP system. The microphone is used by DARE-GP to listen for the smart speaker wake word. Upon hearing the wake word, the EON is played through the EON speaker touching the smart speaker.}
\vskip -2ex
\label{fig:threat}
\end{figure}

While there are multiple state-of-the-art (SotA) audio evasion approaches, all of them are lacking in one or more of the areas discussed above. Details comparing DARE-GP to SotA audio evasion techniques are provided in Section \ref{section:related}. \textbf{\emph{DARE-GP’s ability to simultaneously satisfy all of these attributes makes it the first approach to be deployable in a real-world environment.}} Figure \ref{fig:threat} shows this deployment scenario where a DARE-GP device is used in tandem with a smart speaker to play a pre-generated EON acoustically in real-time.

\textbf{Contributions:} DARE-GP is a first-of-its-kind work to deceive previously unseen black-box SER classifiers while preserving the transcription-relevant content of the speech (Section \ref{eval-online-1}). Unlike previous GP-based audio evasion works \cite{li2020practical, alzantot2018did, wu2019semi, taori2019targeted}, DARE-GP: a) performed \emph{constrained optimization} to protect audio transcription during evasion, and b) DARE-GP used GP to generate \emph{Universal} Adversarial Audio Perturbations (UAAPs) \cite{abdoli2019universal} that can be played as additive noise that causes misclassification of previously unheard utterances in real-time. Further, by leveraging the \emph{strong connection between spectral components of speech and emotional content} (backed by discussed above literature), DARE-GP evasions transfer to \emph{previously unseen black box models} without requiring query access to the models, and without any a priori knowledge of the models' features, topologies, or class labels.  Extensive evaluations demonstrate DARE-GP's: superior performance against state-of-the-art SER evasion techniques (Section \ref{section:evalcompare}), robustness versus a knowledgeable adversary (Section \ref{section:defenses}), and performance in a real-world, over-air deployment scenario, in terms of both SER evasion and transcription utility protection using two off-the-shelf smart speakers. (Section \ref{real-alexa-evaluation}).

\par

The remainder of this paper is organized as follows: Section \ref{section:threats} - Attack Model, Section \ref{section:problemstatement} - Problem Statement, Section \ref{section:techapproach} - Technical Approach, 
 Section \ref{section:datasets} - Datasets, Section \ref{section:evaluation} - Evaluation, Section \ref{section:discussion} - Discussion, Section \ref{section:related} - Related Works, and Section \ref{section:conclusions} - Conclusion.

\section{Attack Model}\label{section:threats}
The following sections provide the context under which DARE-GP was developed, deployed, and evaluated.

\paragraph{\textbf{$\bullet$ Attack Goal: }}

DARE-GP's goal is to degrade inference of a user’s emotional state based upon the spectral components of that user’s interactions with a smart speaker's voice assistant (VA). This obfuscation of the user's emotional state should not come at the expense of usability of the smart speaker's VA; the solution must: (1) minimize transcription errors, and (2) support real-time, on-demand use.

\paragraph{\textbf{$\bullet$ Target System:}}

A black-box SER classifier operated by the smart-speaker VA provider, whose purpose is to infer a user’s emotional state based upon the spectral components (not the transcript) of a user’s speech. 

\paragraph{\textbf{$\bullet$ Proposed System’s Access Level:}}

DARE-GP did not require any privileged access to the smart speaker; it could not rely upon any smart speaker side-channels, undocumented hardware features, or custom hardware/firmware updates. Further, the initial audio processing on the smart speaker was treated as a `closed' system; any custom code/skills on the smart speaker could not preempt upload of the audio to the cloud. In addition, DARE-GP made no assumptions about the black-box SER classifier; DARE-GP’s interactions with the SER classifier were limited to the same audio channel used for user command processing. DARE-GP could not rely upon any access to the black-box SER classifier to train a surrogate classifier \cite{ilyas2018black,cheng2018query,cheng2019sign}. In addition, DARE-GP could not assume a specific feature representations, topology or set of class labels. 

\paragraph{\textbf{$\bullet$ Knowledgeable Adversary:}}
DARE-GP was developed considering that an SER provider would be considered an adversary who wanted to ensure that the SER classifier's emotion inference was correct. As the SER provider, this adversary would have access to audio samples mixed with an EON but not access to the user's unmodified speech, nor would they have access to ground truth labels for the user's speech. This is important because, without this level of access, the SER provider would not be able to use adversarial training \cite{goodfellow2014explaining} to defend against DARE-GP. Further, EONs are different for different households, i.e., targeted set of users (see Section \ref{section:techapproach} and Section \ref{section:discussion}), we assumed that the adversary did not have access to the precise parameters of the EON. DARE-GP's robustness against various SotA audio evasion defenses are presented in Section \ref{section:defenses}. 

\paragraph{\textbf{$\bullet$ Assumptions: }}
The primary assumption was that the SER classifier made emotion inference based upon the spectral information of speech. There were two reasons for this assumption. First, if a user’s interaction transcript did contain emotion information, there is no solution without interfering with the smart speaker’s primary functionality: execution of commands. Secondly, smart-speaker command transcripts do not typically contain emotion information \cite{ammari2019music}; Appendix \ref{appendix:sentiment} contains the results of an experiment that validate this assumption. 

Another important assumption during DARE-GP development related to the speech-to-text implementation used by the smart speaker. Since many smart speakers do not provide APIs to access command transcripts, DARE-GP could not rely upon the smart speaker’s speech-to-text utilities. DARE-GP was developed using an open-source vosk/kaldi \cite{povey2011kaldi} transcription service which we assumed would be less tolerant of noise than the commercially-supported speech-to-text implementations within smart speakers. This assumption proved to be true, as DARE-GP was significantly more successful when evaluated against real smart speakers (Section \ref{real-alexa-evaluation}).

\paragraph{\textbf{$\bullet$ Outcome of the Attack: }}\label{attackmodel-outcome}

The desired outcome of DARE-GP’s attack was degraded SER prediction accuracy of the user’s smart-speaker interactions while preserving transcription accuracy. Further, this outcome should be robust versus defenses deployed by a knowledgeable SER Provider. 

\paragraph{\textbf{$\bullet$ Non-Invasivenes: }}

DARE-GP uses additive noise to degrade SER inference; such perturbations had the potential to be invasive if they were too loud, or if the frequencies used had other negative consequences. For example, frequencies outside of human hearing have been used in previous audio evasion works \cite{zhang2017dolphinattack}; regular exposure to frequencies beyond the range of human hearing can cause health-related side effects \cite{leighton2016some}. To address this, DARE-GP’s audio perturbations were constrained to typical speech frequency ranges and were constrained on amplitude to prevent impacts on the runtime environment. 

\section{Problem Statement}\label{section:problemstatement}

\begin{flushleft}
To formalize the research questions described in the introduction:
\end{flushleft}

\textbf{Given}: A smart speaker executing speech-to-text transcription system \textbf{S} and a black-box SER classifier \textbf{$C$}, within environment \textbf{E}. As is a common practice, the SER classifier \textbf{$C$} has been trained to improve classification performance on user population \textbf{U} \cite{ta2022improving}.

\textbf{Generate}: A single Emotion Obfuscating Noise (EON), \textbf{P}, that increases the misclassification rate of \textbf{C} on utterances generated by users in \textbf{U} within the confines of \textbf{E} without impacting the effectiveness of \textbf{S}.

\begin{flushleft}

\par 
Definitions of some important terms in this section are provided in Table \ref{table:terms}. The \textbf{Technical Approach} (Section \ref{section:techapproach}) and \textbf{Evaluation} (Section \ref{section:evaluation}) address the following challenges associated with answering the research questions:

\end{flushleft}

\begin{itemize}[leftmargin=0.4cm]

\item Implement a constrained optimization approach that, given a surrogate SER classifier \textbf{C*} and a digital environment \textbf{$\hat{E}$}, develops an EON \textbf{P\^} that does not degrade the effectiveness of \textbf{S} (constrain) and increases misclassification in \textbf{C*} that transfers to black-box SER classifiers \textbf{C} (optimization objective). Details of the approach to address this are provided in Section \ref{section:techapproach} and empirical evaluation demonstrating the effectiveness of the generated EON is presented in Section \ref{eval-online-1}.

\item Given a \textit{SER Provider} with knowledge of DARE-GP, assess their ability to infer the true class of user utterances perturbed by \textbf{P\^}. Details regarding these defenses and their effectiveness are provided in Section \ref{section:defenses}.

\item Assess the efficacy of a digitally-developed EON \textbf{P\^} when played in the original real environment \textbf{E}; that is, assess \textbf{P\^} $\rightarrow$ \textbf{P}. Extensive evaluations are provided in Section \ref{real-alexa-evaluation}.

\end{itemize}

\begin{table*}[]
\caption{Definitions of Terms}
\resizebox{0.99\linewidth}{!}{
\begin{tabular}{|p{5cm}|p{16cm}|}
\hline
SER Provider    & Stakeholder responsible for operating and training the SER classifier; the purpose of DARE-GP is to prevent this principle's SER classifier from performing accurate inferences about a user's emotions. \\ \hline
Audio Sample    & Speech sample with a valid transcript and classification by a target emotion classifier. These are the samples for which we want to create evasive variants. \\ \hline
Evasive Audio Sample & Modified version of an audio sample that a) the target classifier cannot determine the actual class and b) the transcript is still correct. \\ \hline
Emotion Obfuscating Noise (EON) &
A combination of $N$ tones that can be combined with an audio sample to generate a potentially evasive sample. \\ \hline
Tone & An acoustic signal specified by a single frequency, offset, duration and amplitude. \\ \hline
Generation (of GP) & A single iteration of a genetic program (GP).\\ \hline
Population & Collection of individuals resulting from a generation.\\ \hline
Individual & A member of the population whose parameters can be used to generate an Emotion Obfuscating Noise (EON).\\ \hline

\end{tabular}
}
\label{table:terms}
\end{table*}

\section{Technical Approach}\label{section:techapproach}

\textbf{EON Description} DARE-GP uses GP to generate an EON. Sound waveforms have three basic physical attributes: frequency, amplitude, and temporal variation \cite{dobie2004basics}. Following the natural sound characteristic, each EON is a mixture of $N$ different tones, where each tone has a different frequency, corresponding amplitude, and temporal variation through different start-time and duration (Table \ref{table:terms}). Since EONs are hearable sound, they are \textit{additive} in nature; meaning can be played simultaneously with the target users’ $U$’s speech in real-time to inject noise. One of the most important characteristics of EONs is that they are \textit{universal} within the set of users \textbf{$U$}. A single EON can be used to alter the utterances from the set of users; separate EONs are not required for new utterances. The number of tones $N$ of EON is a hyperparameter derived empirically in our evaluation.

\textbf{How Does an EON Cause Misclassification?} To explain how an EON cause misclassification we leverage a machine learning prediction explanation technique: Kernel SHAP \cite{lundberg2017unified}. Kernel SHAP approximates the conditional expectations of Shapley values (i.e., the importance of input attributes) in deep learning models by perturbing different portions of the input and observing the impact on output classes. Since we are interested in the spectral attributes of speech that depict emotional information to an SER classifier, the Shapley values are calculated by masking specific frequency bands to measure significance. 

\begin{figure}
\begin{center}
\begin{tabular}{c|c|c}
\includegraphics[height=1.25in,trim={8cm 0 9cm 0},clip]{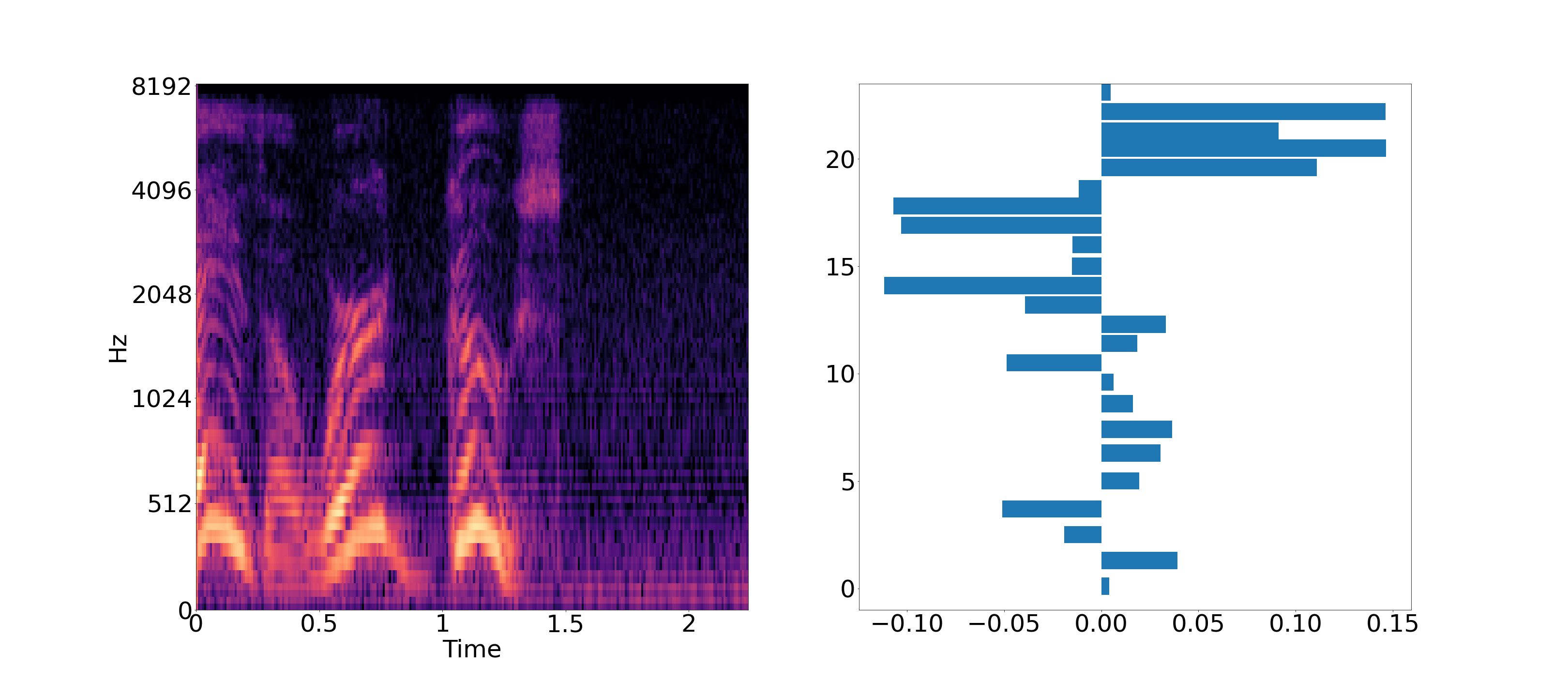} & 
\includegraphics[height=1.25in,trim={1.8cm 0 6cm 0},clip]{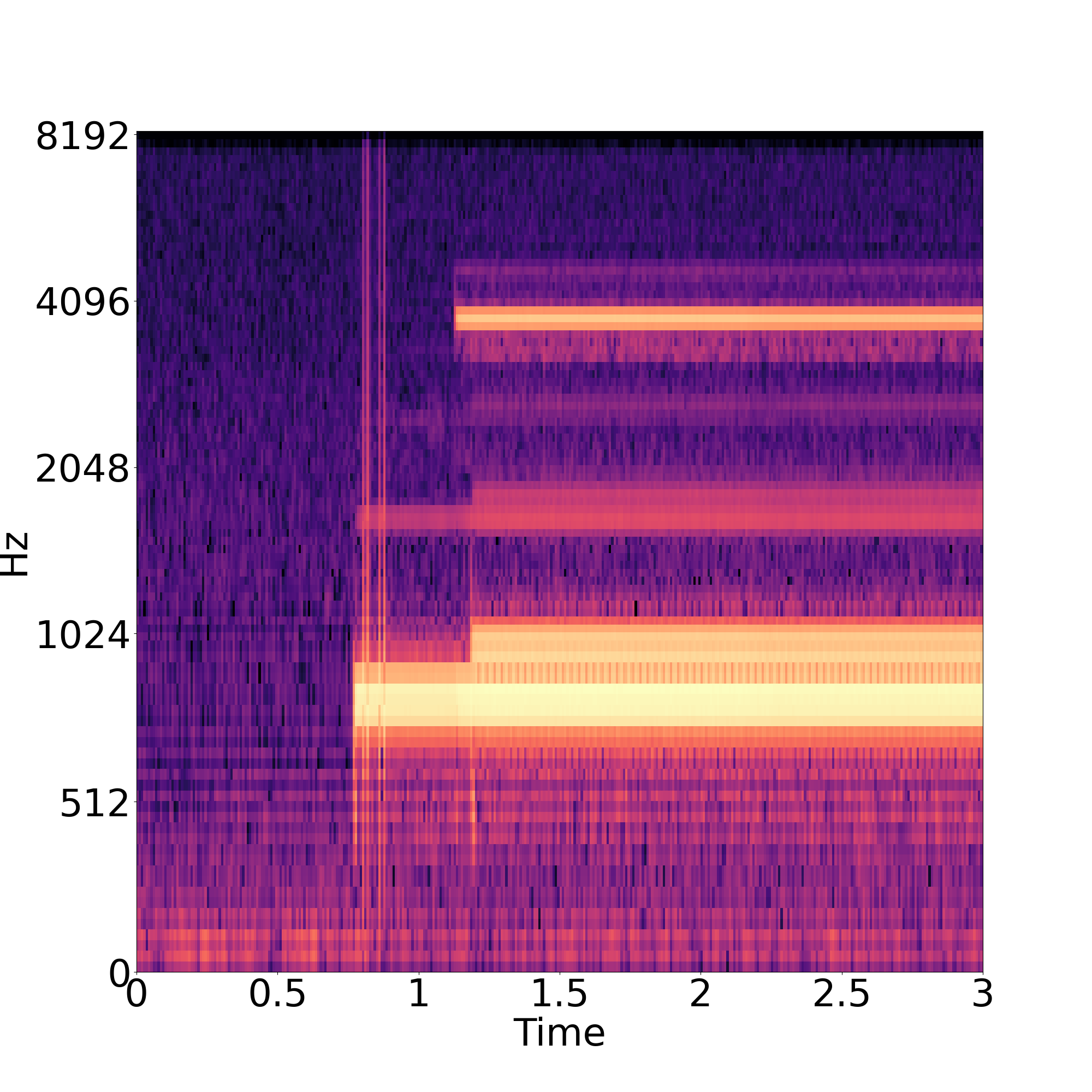} &
\includegraphics[height=1.25in,trim={9.3cm 0 8cm 0},clip]{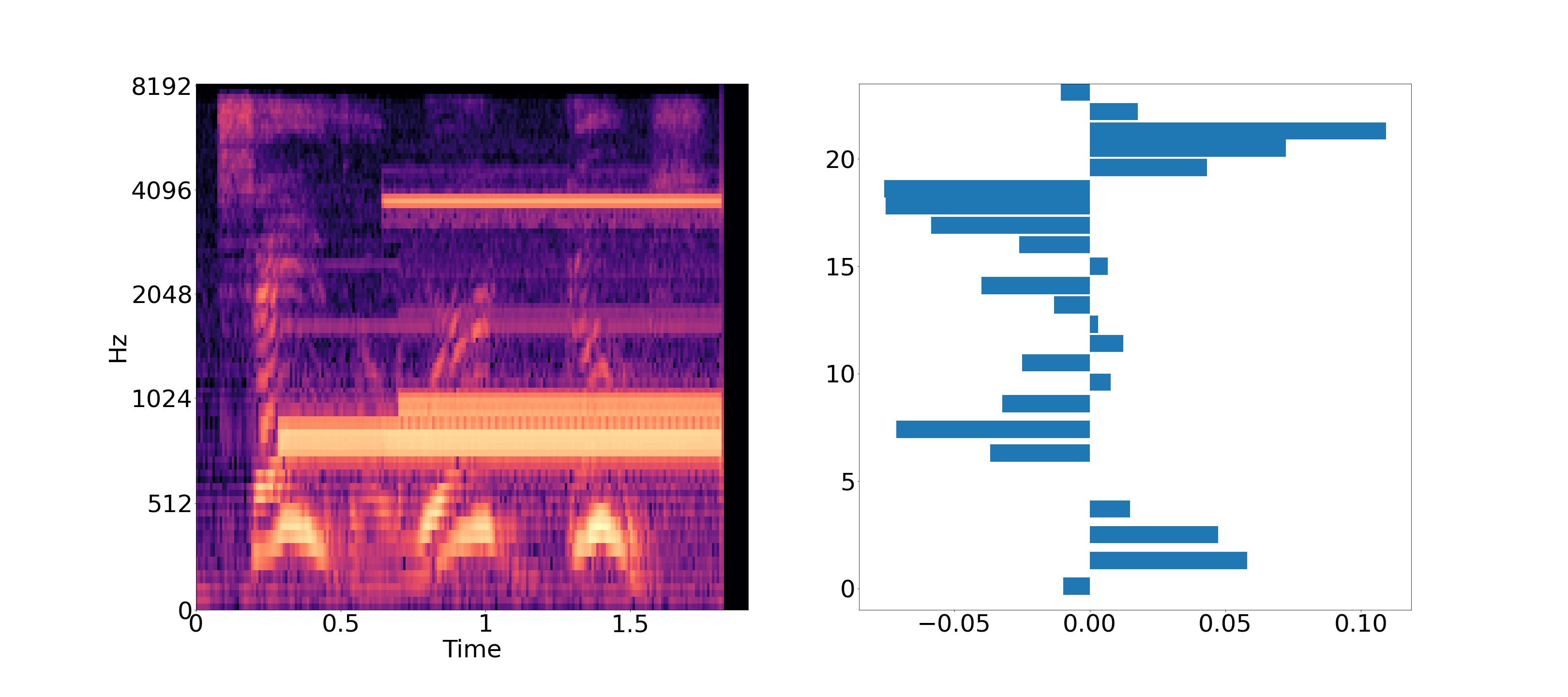} \\

\textbf{SER Prediction:} Angry & Spectral Perturbation  & \textbf{SER Prediction:} Happy\\
\textbf{A} & \textbf{B} & \textbf{C} \\

\end{tabular}
\end{center}
\caption{(A) Spectrogram of a user utterance recorded at a speaker distance of 1ft from Echo DOT and associated Kernel Shap explanation. Correctly classified as \textit{angry}. (B) Spectrogram of a spectral perturbation. (C) Spectrogram of user utterance when played simultaneously with the spectral perturbation and recorded by Echo DOT. The resulting Kernel Shap explanation confirms that the spectral perturbation led to the misclassification of this sample as \textit{happy}.  } 
\label{fig:newshap}
\vskip -1ex
\end{figure}

Figure \ref{fig:newshap} visualizes the spectral attributes on which an SER classifier focuses to make an inference and how DARE-GP spectral perturbation inject noise on those attributes to cause misclassification. Figure \ref{fig:newshap}A is the spectrogram for a user utterance, and the associated Kernel SHAP explanation \cite{lundberg2017unified} for the prediction of class label: \textit{angry}. The blue bars to the right and left indicate frequency bands that contribute the correct and incorrect label, respectively. In this instance, the SER classifier correctly predicted a class label of \textit{angry}. According to the explanation, the frequency bands corresponding to the second (F2) and fourth (F4) formants and high-frequency information (above 4k Hz) are important for correct emotion prediction, which is in line with literature \cite{sondhi2015vocal,erickson2008cross}. 

Figure \ref{fig:newshap}B is the spectrogram of a DARE-GP-generated spectral perturbation composed of three tones. Each tone represents a frequency, an amplitude, and a duration, which injects noise in a specific frequency band of the target set of users' speech through which they commonly depict emotion. The number of tones is a hyperparameter that we determined through empirical analysis. Each tone's frequency, amplitude, and duration are learnable parameters that we learned through GP. Notably, Figure \ref{fig:newshap}B's tones correspond to the above-discussed important frequency bands for SER’s correct classification. 

Finally, Figure \ref{fig:newshap}C shows the spectrogram of the user utterance when played simultaneously with the spectral perturbation and again presented to the SER classifier. Notably, according to the Kernel SHAP explanation regarding the \textit{angry} class, the frequency bands affected by the spectral perturbation strongly correlate with misclassification. These frequency bands were positively impacting SER’s correct \textit{angry} class prediction (right-side bars in Figure \ref{fig:newshap}A); however, due to the injection of noise, they are negatively impacting SER’s angry class prediction (changed to left-side bars in Figure \ref{fig:newshap}C), causing misclassification to \textit{happy}.

\textbf{Use of Genetic Programming (GP):} 
Using constrained GP, DARE-GP can develop an EON for a target set of users that causes misclassification while constraining them to limit transcription errors. This is done by incorporating both evasion and transcription accuracy into the fitness function used in the GP evolution process. Most importantly, EON development does not rely on the specifics of the feature representation used by a target SER classifier. Rather than trying to replicate a specific SER classifier’s gradients, the GP fitness function guides the evolution of perturbations that mask frequency bands through which a given set of users depict their emotional information. By injecting noise in a target users' emotion-relevant spectral traits the EON a) transfers to a diverse set of previously unseen black-box SER classifiers, b) is universal, i.e., effective on previously unseen new utterance, and c) functions well acoustically.

\begin{figure*}[t]
\centering
\includegraphics[width=0.7\textwidth]{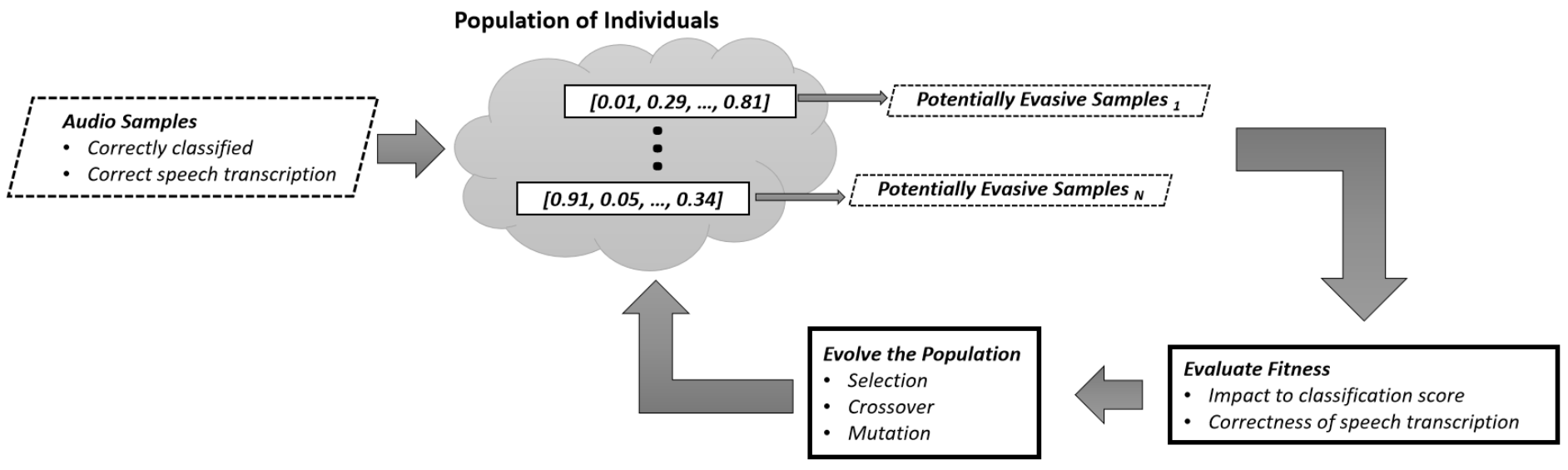}
\vskip -2ex
\caption{Genetic Programming Adversarial Evasion Workflow}
\label{fig:workflow}
\vskip -1ex
\end{figure*}

\textbf{Genetic Programming (GP) Workflow:} Figure \ref{fig:workflow} shows the high-level GP approach used to generate an EON. The approach requires a labeled set of audio samples from the users. There are two constraints on this dataset: samples need to be correctly classified by the surrogate SER classifier, and the transcription service needs to be able to correctly extract text transcripts from these samples. Naturally misclassified samples are not interesting because they are already misclassified. Likewise, audio samples that cannot be transcribed correctly are not useful because they do not support constraining the EONs to limit transcription errors. 
\par 
As the first step of GP workflow in Figure \ref{fig:workflow}, a population of \textit{individuals} (Table \ref{table:terms}), each of which has the parameters necessary to generate an \textit{EON}, is initialized. These individuals are then assigned \textit{fitness scores} by generating their EONs, combining those EONs with audio samples from the training set to generate potentially evasive audio samples, and then assessing the fitness of those samples with respect to the classifier \textbf{C*} and the utility of those samples with respect to the transcription system \textbf{S}. 
A new set of individuals, i.e., a new population is generated by \textit{selecting} the strongest (i.e., higher fitness) individuals and
creating new variations from them through \textit{crossover} and \textit{mutation}. This process is iterated over multiple generations (i.e., GP iterations), with each generation’s individuals generating slightly better EONs (i.e., having higher fitness).

Specific details of the GP operations are provided below:

\begin{itemize}[leftmargin=0.0cm]
\item[] \textbf{Fitness -} The fitness calculation ranks each individual in the population. For DARE-GP, this ranking is based upon the ability of an EON to mislead the surrogate SER classifier \textbf{C*}, and its ability to do so without compromising the underlying audio's transcription (see Equations \ref{eq:deception}-\ref{eq:fitness}).

\item[] \textbf{Selection -} This step selects a subset of individuals from a population to carry forward into the next generation before crossover and mutation. Selection is performed using a tournament selection method \cite{koza2005genetic} with a tournament size of $nSel$. This guaranteed that at least the $nSel - 1$ weakest individuals were eliminated from each generation. 

\item[] \textbf{Crossover -} In GP crossover, offspring “…are created by exchanging the genes of parents among themselves” \cite{Mallawaarachchi2017}. This step is a method to generate new individuals (i.e., offspring) from previous selected ones (i.e., parents), thus generating new EONs, by combining the parameters of two existing individuals to create two new individuals. Crossover is performed by exchanging EON-parameters between the selected parents \cite{Mallawaarachchi2017} with probability  $pCX$.

\item[] \textbf{Mutation -} Used to prevent population stagnation \cite{rajabi2021stagnation}. New individuals are generated by randomly modifying select individuals' EON-generation-parameters. Mutation introduces the greatest amount of variability in the population; a given individual can undergo any number of changes, leading to significant improvement or degradation. It is performed by randomly shuffling EON parameters (scaled to [0,1]) with probability $pMUX$.

\item[] \textbf{Final EON Selection -} After iterating for $K$ generations, the EON with the highest \textit{evasion success rate (ESR)} when mixed with the validation dataset. 

\item[] \textbf{Hyperparameters} - $nSel$, $pCX$, and $pMUX$ are hyperparameters, identified empirically through grid-search.

\end{itemize}

Important questions pertaining to the approach are answered below:
\paragraph{$\bullet$ \textit{\textbf{Are EONs developed with access to the black-box SER classifiers?}}}

\par
No. The development of EONs exclusively leverages \textbf{C*}, a surrogate SER classifier created for this purpose. It is important to emphasize that, while the topology and weights of this classifier are available during EON development, the only privileged characteristics of \textbf{C*} used are the scores from its softmax layer during fitness evaluation. 

\paragraph{$\bullet$ \textit{\textbf{How are EONs constrained to prevent degraded transcription?}}}

\par
As mentioned previously, DARE-GP addresses a constrained optimization problem where SER classifier deception needs to balance against audio utility. The balance is maintained by incorporating a transcription correctness score (Equation \ref{eq:transcription}) into the EON fitness function (Equation \ref{eq:fitness}). DARE-GP used an open-source audio transcription library backed with a kaldi \cite{povey2011kaldi} speech model to calculate the transcription correctness score. DARE-GP was developed under the assumption that a commercial transcription solution would be at least as robust to background noise as the open-source implementation used for these experiments. This assumption was validated with the experiments in Section \ref{real-alexa-evaluation}.

\paragraph{$\bullet$ \textit{\textbf{How is the EON fitness calculated?}}} The fitness score measures the relative efficacy of individuals in a population and is calculated using a fitness function. For DARE-GP, this fitness function is composed of two components: a \textit{deception score} and a \textit{transcription score}. The deception score, given in Equation \ref{eq:deception}, measures the extent to which the EON generated by an individual fools the surrogate classifier \textbf{C*}. 

\begin{equation} \label{eq:deception}
\resizebox{0.7\linewidth}{!}{
\begin{math}
  \begin{aligned}
	deception(ind) = \frac{\sum_{x \in S} max(0, [smax(x, c_{true}) - smax(x \oplus ind, c_{true})]) }{|S|} + \sum_{x \in S} bonus(ind, x) \\
	 bonus(ind, x) = \begin{cases} 
  b, & if \exists_{c_{other} \neq c_{true}} : smax(x \oplus ind, c_{true}) < smax(x \oplus ind, c_{other})\\
  0, & otherwise
 \end{cases}
	\end{aligned}
 \end{math}
}
 \end{equation}

where:
\\

$S \overset{def}{=} $a random subset of the audio training data

$b \overset{def}{=} $a bonus awarded  for each successful misclassification

$c_{true} \overset{def}{=} $the actual class for x

$c_{other} \overset{def}{=} $any class other than $c_{true}$

$smax(x,c) \overset{def}{=} $the value of class c in the softmax result from simple\_classifier.predict(x)

$x  \oplus ind \overset{def}{=}$the result of mixing the audio sample x with the EON ind
\\

The first term of the deception score is the mean decrease in the actual class’ score from the softmax in our surrogate classifier. The second term is a bonus applied for each misclassification; the purpose of this term is to prioritize increasing the number of misclassifications rather than increasing the confidence of existing misclassifications. The bonus value used in this work was $50$ and was found empirically using a grid search. The transcription score in Equation \ref{eq:transcription} penalizes transcription errors:

\begin{equation}
\resizebox{0.7\linewidth}{!}{
$transcription(ind) = 
 \begin{cases} 
  0, & if \exists_{x \in S} : tscr(x \oplus ind) \neq tscr(x) \\
  0, & if \exists_{x \in S} : conf(x \oplus ind) < t \\
  1 - \frac{\sum_{x \in S} max([conf(x) - conf(x \oplus ind)], 0)}{|S|}, & otherwise
 \end{cases}$
 }
\label{eq:transcription}
\end{equation}
where:

$tscr(x) \overset{def}{=}$ the transcript extracted from audio sample x

$conf(x)\overset{def}{=}$ the confidence of the transcription result for x, scaled to [0,1]

$t \overset{def}{=}$ the minimum acceptable transcription confidence.\\

The transcription score goes to $0$ if any of the audio samples when mixed with an individual’s EON, cannot be transcribed correctly. In addition, if any of the transcription confidence scores fall below a specified confidence threshold, then the transcription score goes to $0$. Otherwise, the score is $1$ minus the mean confidence decrease as a result of applying the EON to audio samples in S. Confidence increases never occurred during the experiments for this work, but to accommodate this possibility a confidence increase would be replaced by $0$ when calculating this mean. The fitness score is calculated as the product of the deception(ind) which is a real number, and the transcription(ind) which is a real number in [$0$, $1$], where a higher value indicates better preservation of the audio transcripts.

\begin{equation}
fitness(ind) = deception(ind) \times transcription(ind)
\label{eq:fitness}
\end{equation}

Taking the product of the deception and transcription scores allows for incremental improvements to classifier deception while harshly penalizing transcription errors. This approach was taken due to an obvious complication; a sufficiently loud EON could drown out the speech in the audio samples, thus providing near-perfect evasive quality while rendering the audio effectively useless.

This fitness function exclusively uses the surrogate SER classifier \textbf{C*}. In our evaluation, the surrogate SER classifier is a simple DNN that demonstrates state-of-the-art performance using Mel-frequency cepstral coefficients (MFCCs) as the feature representation; MFCCs are a widely used feature in many audio classification use cases \cite{wu2006multiple,logan2000mel,blum1999method,cowling2003comparison,rabaoui2007towards,eronen2005audio}. Additional details of this classifier are provided in Table \ref{table:classifiers}. 

\paragraph{$\bullet$ \textit{\textbf{How will DARE-GP generate an EON for a set of users in a real environment?}}} As shown in Figure \ref{fig:threat}, the DARE-GP system will include a microphone that listens for a wake word and 
a speaker that 
will physically touch the smart speaker and play the `final EON' once the wake word is heard. By touching the smart speaker, the DARE-GP speaker can play the EON with minimal loudness (i.e., minimally invasive) since the EON acoustic will propagate both via air and physical (i.e., smart speaker material) modalities.
\par
\begin{figure}[t]
\centering
\includegraphics[width=0.8\linewidth]{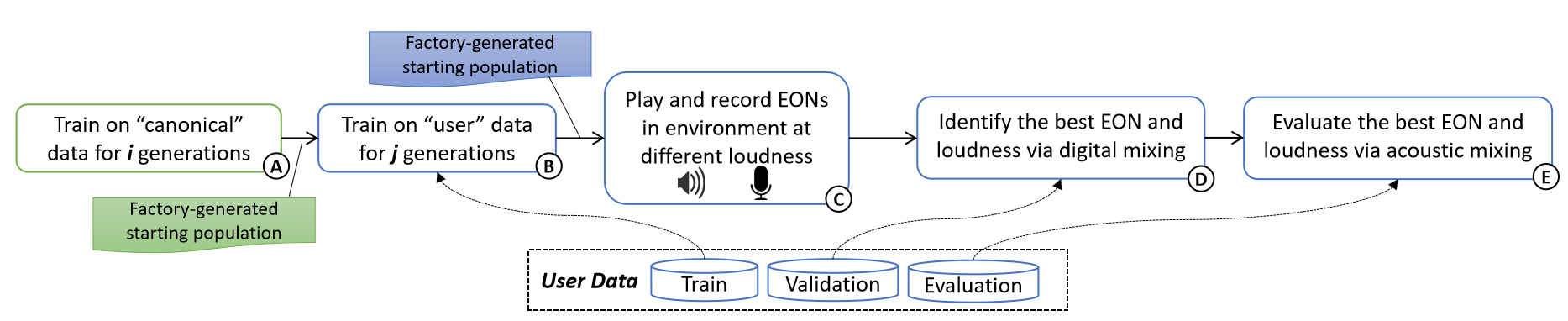}
\vskip -1ex
\caption { (A) Pre-train a generic population  using the “canonical” dataset.  (B) Starting with this pre-trained population, generate a tailored population for the targeted users by training with a subset of  the “user” dataset.  (C) Play the EONs generated by this tailored population at varying loudness in environment E and record them. (D)  Pick the top EON/loudness combinations by digitally mixing with the “user” validation data and calculating evasiveness for each EON/loudness combination. (E) Over-air acoustically mix the best EON/loudness with the “user” evaluation dataset and calculate evasiveness.}
\label{fig:structure}
\end{figure}

 
The real-world deployment process is presented in Figure \ref{fig:structure}.
As the initial step (step A), DARE-GP will digitally train/develop a set of EONs on existing independent datasets (i.e., canonical data). These are ``factory default’’ generic EONs without any tailoring for the target users \textbf{U} and target environment \textbf{E}. By initializing these EONs, the amount of in-home fine-tuning is significantly reduced. To adapt the final single EON for the target users \textbf{U}, DARE-GP will record some speech samples from them in the target environment \textbf{E} (details provided in Appendix \ref{appendix:acousticrecordings}) and digitally fine-tune the ``factory default’’ generic EONs to the users’ speech through ‘$j$’ iterations/generations of the GP (step B). The final generation of GP will result in ‘$t$’ EONs with the highest fitness in the digital environment. However, to identify which of the final ‘t’ EONs perform best over-air for the environment \textbf{E}, DARE-GP plays and records the ‘$t$’ EONs with different loudnesses (step C) and again digitally evaluates their fitness with the users’ \textbf{U}’s recorded speech. Finally (step D), the highest fitness EON-and-loudness pair out of all possible candidates is selected as the \textbf{EON} for the target users \textbf{U} to deploy for the target environment \textbf{E}. Step E is the evaluation step of the final EON by simultaneously playing it with users' speech (i.e., acoustic mixing). As discussed above, the EON to propagates via conductance \cite{weik2012communications}; since the EON is not presented to the smart speaker microphone from only one direction, the built-in beamforming is not able to single out the EON (i.e., noise).

\section{Datasets}\label{section:datasets}
The experiments described in this paper were performed using two standard audio datasets: the Ryerson Audio-Visual Database of Emotional Speech and Song (RAVDESS) \cite{livingstone2018ryerson}, and the Toronto Emotional Speech Set (TESS) \cite{dupuis2010toronto} datasets. Details of both of these datasets are provided in Table \ref{table:dataset}.

The spoken part of the RAVDESS dataset (no sung content) consists of $1,440$ samples from $24$ speakers. The speakers were actors, split evenly between males and females. The speakers performed two separate utterances while demonstrating $8$ different emotions. The subset of the TESS dataset used in this paper consisted of $1800$ audio samples generated by two actresses. These samples spanned five emotions (neutral, angry, happy, sad, and fearful) and included $200$ utterances. This dataset was utilized in assessing the extensibility of an EON to multiple, previously unheard utterances.

The counts of audio samples for these datasets are provided in Table \ref{table:dataset}, but not all of these samples were usable for the experiments in this work. Before performing any experiments, specific samples were removed if a) the speech-to-text library was unable to extract a correct transcript or b) the surrogate SER classifier misclassified the sample before any EONs were applied. This process ensured that EONs were not penalized for audio samples that did not normally transcribe correctly, and evasion scores were not inflated due to natural misclassifications from the surrogate classifier. 

Two additional datasets were implicitly used in this work: Interactive Emotional Dyadic Motion Capture (IEMOCAP) \cite{busso2008iemocap} and Surrey Audio-Visual Expressed Emotion (SAVEE) \cite{jackson2014surrey}. These datasets were part of the training datasets for several of the black-box SER models used in the evaluations of DARE-GP. These datasets were not used during EON development or evaluation. 
\vspace{0.5cm}

\begin{minipage}[c]{0.5\textwidth}
\captionof{table}{Dataset details}
\resizebox{0.9\linewidth}{!}{
\begin{tabular}{|c|c|c|c|c|}
\hline
 \textbf{Dataset} & \textbf{Speakers} & \makecell[c]{\textbf{Distinct}\\ 
 \textbf{Utterances}} & \makecell[c]{\textbf{\# of} \\\textbf{Classes}} & \makecell[c]{\textbf{Total} \\ \textbf{Recordings}} \\ \hline
 RAVDESS \cite{livingstone2018ryerson}& 24 & 2 & 8 & 1440 \\ \hline
 TESS \cite{dupuis2010toronto}& 2 & 200 & 5 &1800 \\ \hline
\end{tabular}
}
\label{table:dataset}
\end{minipage}
\begin{minipage}[c]{0.5\textwidth}
    \centering
\captionof{table}{EON Generation Meta Parameter Values}
\resizebox{0.9\linewidth}{!}{
\begin{tabular}{|l|c|l|c|}
\hline
\textbf{Parameter} & \textbf{Value} &\textbf{Parameter} & \textbf{Value} \\ \hline
\makecell[l]{Number of Tones\\in an EON}& $3$ &EON Frequency  & [$100$,$4000$] Hz\\ \hline
\makecell[l]{Tournament Selection\\Size ($nSel$)} & $3$ &EON Duration   & [$2.5$, $4.0$] s\\ \hline
\makecell[l]{Crossover Probability\\($pCX$)} & $0.5$ &EON Offset   & [$0.0$, $0.5$] s\\ \hline
\makecell[l]{Mutation Probability\\($pMUT$)}  & $0.1$ &EON Amplitude & [$0.0067$, $0.04$]\\ \hline
\end{tabular}
}
\label{table:metaparams}
\end{minipage}

\section{Evaluation}\label{section:evaluation}
The following experiments address the following \textbf{research questions}:

\begin{enumerate}

    \item \textit{Given a set of users, can an approach deceive a (i) previously unseen black-box SER classifier (ii) without compromising the speech-to-text transcription on (iii) previously unheard utterances?} (Section \ref{eval-online-1})    

    \item \textit{How does the performance of this approach compare with SotA audio evasion techniques?} (Section \ref{section:evalcompare})

    \item \textit{Can a knowledgeable SER operator defend against this technique?} (Section \ref{section:defenses})

\end{enumerate}

\begin{flushleft}
The previous research questions were evaluated digitally; user utterances and EONs were mixed in code. The following question considers the real-world over-air deployment of digitally-generated EONs.
\end{flushleft}

\begin{enumerate}

\setcounter{enumi}{3}

    \item \textit{Can such protection be granted in an acoustic, real-world scenario with: closed, off-the-shelf (OTS) smart speakers, variable user location related to the smart speaker, and SWAP \cite{spitzer2017rtca} constraints?} (Section \ref{real-alexa-evaluation})
    
\end{enumerate}

To evaluate the success of DARE-GP at answering these questions, the following metrics were used:

\begin{itemize}[leftmargin=0.2cm]
  \item[] \textbf{Evasion Success Rate \emph{(ESR)}:} The fraction of evaluation samples that both fool the target SER classifier and transcribe correctly. This metric ensures that a solution both protects emotional privacy \textbf{and} utility of the audio modified by the solution.

  \item[] \textbf{False Label Independence:} Relationship between an utterance's actual emotion label and the fake label caused by DARE-GP. For example, if \textit{calm} audio samples, when perturbed, always resulted in samples with the false emotion \textit{happy}, it would be trivial to discern the true emotion. This metric assesses the strength of the relationship between the true and the false emotion/class. Two information theoretic measures are used for this metric: \textit{\textbf{Normalized Mutual Information (NMI)}} \cite{estevez2009normalized} and \textit{\textbf{Matthews Correlation Coefficient (MCC)}} \cite{chicco2020advantages}.
\end{itemize}

\par
\begin{figure}[]
\centering
\includegraphics[width=0.55\linewidth,clip]{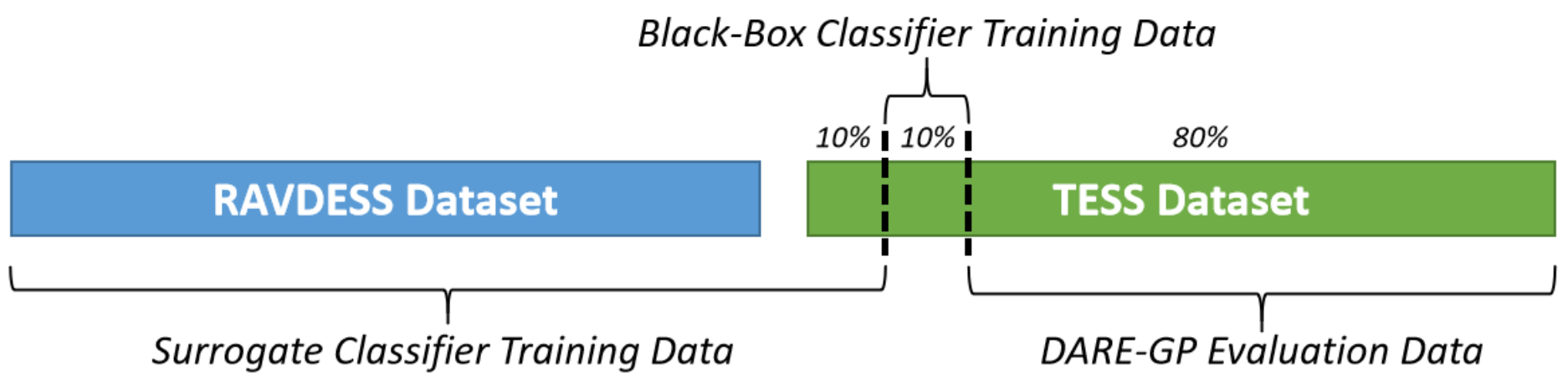}
\vskip -3ex
\caption{A high-level view of how the datasets from Section \ref{section:datasets} were used in these evaluations. All RAVDESS data was used for EON training, as was a 10\% slice of TESS data. An additional 10\% of TESS data was reserved for additional training on any black-box classifiers that underperformed on the original TESS data. The remaining 80\% of TESS data was used for evaluation. All TESS data splits were balanced between the two speakers in the dataset.}
\label{fig:split}
\end{figure}

\textbf{Dataset Split for Black Box Evaluation:}
All evaluations performed in the following sections used EONs that were trained on RAVDESS data for 40 generations, tailored on 10\% of the TESS data for 10 generations, and evaluated using the 80\% of TESS data dedicated for evaluation (see Figure \ref{fig:split}). RAVDESS was used to train "factory default" EONs without any insight into the target environment or users. The TESS data (the portion used to train \& select the EON) acted as user-supplied audio samples to tailor the "factory default" EONs to the target users. 

Notably, the black box SER classifiers that DARE-GP aims to deceive, are taken from git-repositories or trained by us on a third independent dataset, IEMOCAP \cite{busso2008iemocap}. The other 10\% of TESS data was reserved for further training black box SER classifiers, if their performance was too low without any TESS data fine-tuning, which is a known limitation of SER models \cite{ta2022improving}. Such a split ensures, the EON development process, Black box SER training process, and evaluation process use completely disjoint data; prohibiting any data leakage.

All splits on TESS data were balanced with respect to the two speakers in the dataset, and were based upon utterance ID; utterances were disjoint between the EON training, evaluation and SER training sets. Such a split ensures that the evaluations are performed on previously unseen utterances/commands.

\textbf{Metaparameters:} The metaparameter values (discussed in Section \ref{section:techapproach}) are provided in Table \ref{table:metaparams}. These parameters were found empirically based on the desired level of diversity and EON effectiveness. The EON Frequency Range was set to limit EON frequencies to ranges within the typical human speech. The EON amplitude range was determined based on some initial assessments of EON loudness on transcription scores. The range of amplitude values was limited to protect transcription accuracy and prevent EONs from becoming irritating to users. The EON duration and offset were introduced to prevent EONs from being a single, monotonous tone. This was empirically demonstrated to be more effective and would be more difficult to detect. Similar to the perturbations in \cite{li2020advpulse}, EONs did not require precise alignment with user utterances in order to generate adversarial samples. Further refinement of the metaparameters presented here is left as future work.

Most importantly, all metaparameter values were derived \textbf{without} any access to the black-box SER classifiers and these values were invariant across all of the evaluations presented in this work. This consistency prevented the introduction of any hidden bias based on knowledge of the black-box classifiers' inner workings. 

\subsection{Evasion Success and Transferability}\label{eval-online-1}
This section evaluates the effectiveness of the EON developed utilizing a surrogate SER classifier \textbf{C*} and evaluated against (1) \emph{previously-unseen, black-box classifiers}, \textbf{C} on (2) \textit{previously unheard utterances}. As discussed above, the EON was not tailored for the black-box classifiers in any way. The details of these black-box SER classifiers are provided in Table \ref{table:classifiers}. The measure of effectiveness used was ESR (Section \ref{attackmodel-outcome}). 

\begin{table*}[]
\caption{Classifiers for Transferability Evaluation.}
\resizebox{0.6\linewidth}{!}{
\begin{tabular}{|l|l|l|c|c|}
\hline
\textbf{Classifier}& \textbf{Topology} & \textbf{Features} & \textbf{Classes} & \textbf{Accuracy (Benign)}\\ \hline
Surrogate DNN & 7-layer DNN & MFCC & 8 & 0.877 \\ \hline
Data Flair \cite{flair2019} & 2-layer DNN & MFCC, Chroma & 4 & 0.665 \\ \hline
\makecell[l]{Speech Emotion Analyzer\\ (SEA) \cite{sea2021}} & 18-layer CNN & MFCC & 10 & 0.921 \\ \hline
\makecell[l]{Speech Emotion Classification\\ (SEC)  \cite{sea2021}} & \makecell[l]{Bi-Dir LSTM\\ and CNN} & Mel Spectrogram & 8 & 0.916 \\ \hline
\makecell[l]{CNN with Multi-scale Area Attention\\ (CNN-MAA) \cite{gao2022black, xu2021speech}} & CNN-MAA & Mel Spectrogram & 8 & 0.952 \\ \hline
\makecell[l]{Time Delay Neural Network\\ (TDNN)} & TDNN & Mel Spectrogram & 8 & 0.678 \\ \hline
\makecell[l]{Residual Neural Network\\ (RESNET)}   & RESNET & Mel Spectrogram & 8 & 0.773 \\ \hline
wav2vec \cite{baevski2020wav2vec} & CNN and DNN & wav2vec & 4 & 0.963\\ \hline

\end{tabular}
}

\label{table:classifiers}
\vskip -4ex
\end{table*}

These classifiers used different combinations of Mel-frequency cepstral coefficients (MFCCs) \cite{mogran2004automatic}, mean Chroma \cite{shepard1964circularity} and mean Mel-scaled spectrogram \cite{stevens1937scale} values for their features. In addition, the class labels and network topologies were different between classifiers. Since DARE-GP is oblivious to these details, application to these other classifiers was completely transparent. The results of applying the EON is presented in Table \ref{table:compare}. It is important to note that the \emph{ESR} was calculated on samples that the classifiers originally classified correctly.

The \emph{ESR} against the SEC, TDNN, wav2vec, and RESNET classifiers were particularly significant; these classifiers used Mel-scaled spectrograms for their feature representation and were the least similar to the surrogate classifier in both topology and feature representation. These results underscore the appeal of DARE-GP; since EON is generated to create spectral noise, it is transferable to \textit{previously unseen SER classifiers and unheard utterances}. 

\subsection{Comparison with Other Evasion Attacks}\label{section:evalcompare}

The evaluations in this section compare DARE-GP with state-of-the-art audio evasion attacks. Recently, Gao et al. \cite{gao2022black} evaded SER classifiers using three spectral envelope attacks: Vocal Tract Length Normalization (VTLN) \cite{lee1998frequency}, the McAdams transformation \cite{mcadams1984spectral}, and Modulation Spectrum Smoothing (MSS) \cite{takamichi2015naist}. To compare gradient-based approach's efficacy for this paper's attack model, evaluations were also performed using two white-box, gradient-based attacks: Fast Gradient Sign Method (FGSM) \cite{goodfellow2014explaining} and Projected Gradient Descent (PGD) \cite{madry2017towards}. It is important to note that none of these attacks are executable in a real-time, acoustic environment (see Section \ref{section:advevasion}); hence are not the true baseline or competitors of this paper. However, during the time of writing, these were the most comparable attacks. The evasion results are provided in Table \ref{table:compare}.

\begin{table}
  \caption{ESR values for various attacks against the black-box SER classifiers. VTLN, McAdams, and MSS are universal adversarial attacks performed on all audio samples proposed by Gao, et. al. \cite{gao2022black}. FGSM and PGD attacks were performed as usual, generating bespoke transformations for each audio sample. Multiple values of $\epsilon$ were evaluated; $\epsilon = 0.01$ was the optimal value for both attacks. The DARE-GP ESR was calculated using an EON generated using the surrogate classifier.}
\resizebox{0.6\linewidth}{!}{
\begin{tabular}{|l|c|c|c|c|c|c|}
\hline
\textbf{Classifier} &	\textbf{VTLN} &	
\textbf{McAdams}&	\textbf{MSS}&	\textbf{FGSM} &	\textbf{PGD} &
\textbf{DARE-GP}\\ \hline
Data Flair&	0.455 &	0.063&	0.0  &	0.0 & 0.0 & \textbf{0.854} \\ \hline
SEA&	    0.230 &	0.071&	0.023&	0.0 & 0.0 & \textbf{0.675} \\ \hline
SEC	&       0.338 &	0.077 &	0.061 &	0.0 & 0.0 & \textbf{0.688} \\ \hline
CNN MAA	&   0.302 &	0.043 &	0.044 &	0.0 & 0.0 & \textbf{0.668} \\ \hline
TDNN	&   0.251 &	0.065 &	0.082 &	0.0 & 0.0 & \textbf{0.69} \\ \hline
RESNET	&   0.327 &	0.082 &	0.003 &	0.0 & 0.0 & \textbf{0.683} \\ \hline
wav2vec &   0.187 & 0.058 & 0.001 & 0.0 & 0.0 & \textbf{0.304}\\ \hline
\end{tabular}
}
  \label{table:compare}
\end{table}

Existing literature \cite{waaramaa2006role,waaramaa2008monopitched} shows that the speech spectral attributes, such as the third and fourth formants convey the emotional content of speech, and the three spectral envelope attacks distort the spectral attributes, including temporal content and the formant information; hence can deceive the SER classifiers. However, these transformations are generic and applied to all frequency ranges and harmonics, distorting speech's transcription utility as well, resulting in a relatively low ESR. Moreover, spectral envelope attacks perform relatively well on SER classifiers that take spectrogram directly as input (i.e., SEC, TDNN, RESNET, and wav2vec). They are less effective on the classifiers that apply MFCC on the acoustic data (i.e., DNN, Flair, SEA, CNN-MAA); this transformation filters out some of the distorted information during the feature extraction phase. In contrast, DARE-GP only introduces additive noise on the specific spectral aspects (through generated tones for specific frequencies and amplitudes), which convey emotional information and does not interrupt the speech transcription-relevant information, nor does it modify the overall spectral attributes; hence outperforming the spectral envelope attacks and more importantly performs well against all SER classifiers. 

These evaluations further demonstrate DARE-GP’s generalizability and efficacy against different SER classifiers, even outperforming the attacks that need off-line spectral transformations \cite{gao2022black}. It is important to note that the gradient based attacks, PGD \cite{madry2017towards} and FGSM \cite{goodfellow2014explaining}, were quite ineffective. Both were able to generate evasive audio samples in feature space for more than 90\% of the evaluation audio samples. However, when converted back to audio time series from feature space, the resulting audio was too garbled to transcribe correctly. These transcription errors are what drove the ESR in each trial to 0 for these attacks. 

\subsection{Defenses against DARE-GP}\label{section:defenses}

DARE-GP is an evasion attack against an unknown SER classifier. This attack's effectiveness has been shown above, but this begs the question regarding whether or not a knowledgeable \textit{SER Provider}, one familiar with the details of DARE-GP, could deploy countermeasures to allow them to perform accurate emotion inference. This question is considered from two different vantage points: \textbf{active} and \textbf{passive} defenses. The datasets and EON used in these evaluations were the same used in Section \ref{section:evalcompare}.

\vspace{-1ex}

\subsubsection{Active Defense}\label{section:defenses2}

\textbf{Active} defenses attempt to mitigate the impacts of DARE-GP before attempting to infer the true class. One common approach, adversarial training, is not viable because the \textit{SER Provider} would not have access to the ground truth emotional content associated with EON-modified audio samples. Table \ref{table:Defenses2} shows the evaluation results of several different active defenses against the DARE-GP.

Two of the defense methods considered were first published in  WaveGuard \cite{hussain2021waveguard}: \textit{Audio Resampling} and \textit{Mel Spectrogram Extraction and Inversion}. These methods can be applied as a pre-processing step on any inputs to the SER classifier. \textit{Audio resampling} was ineffective; the tones used to comprise EONs are constant over their duration and are present regardless of the sample rate. Mel spectrogram extraction and inversion was also ineffective. The surrogate classifier and several of the black-box classifiers use MFCCs as their feature representations; if EONs were significantly degraded during the MFCC extraction process, then they would not have been effective against any of these MFCC-based classifiers. One interesting result here was that several applications of defenses actually \textit{\textbf{improved}} ESR against Data Flair and wav2vec. For example, Data Flair uses $40$ MFCCs as the feature representation; the Mel Spectrogram Defense used $22$ MFCCs to be consistent with \cite{gao2022black}. This lossy preprocessing negatively impacted Data Flair's performance on a few audio samples, which degraded the performance and improved the ESR.

In addition to these defenses, if an SER Provider knew the specifics of DARE-GP, and could further capture the specific frequency bands masked by a specific EON, then it would be possible for the provider to implement an
\textit{EON Band Pass} filter. This is not the same as inverting EON application; since, the SER Provider would not have ground truth with respect to the user's speech before EON application. Bandpass filters are \textit{"a linear transformation of the data that leaves intact the components of the data within a specified band of frequencies and eliminates all other components"}  \cite{christiano2003band}. A bandpass filter is an effective means for removing spurious audio data outside frequency ranges of interest (e.g., those of human speech). In this case, bandpass filters were applied to remove within +/- 10Hz  of the EON frequencies. Again, due to the interspersion of EON frequencies with human speech, removing these bands severely impacted the SER classifiers.

\vspace{.5cm}
\begin{minipage}[c]{.48\textwidth}
\captionsetup{width=0.9\linewidth}
\captionof{table}{The effects of various defenses against DARE-GP on the effective ESR against each SER classifier.                     }
\resizebox{1.0\textwidth}{!}{

\begin{tabular}{|l|c|c|c|c|}
\hline
 & \textbf{ESR with} & \multicolumn{3}{c|}{\textbf {$\Delta$ after applying the defenses}} \\
 &	\textbf{No Defense} &		\textbf{\makecell[c]{Audio \\Resmpling}}	&	\textbf{\makecell[c]{Mel Spectrogram \\Extraction}}&	\textbf{\makecell[c]{EON Band\\ Pass}}\\ \hline

\textbf{Data Flair}&	0.839 &		0.000	&	+0.011 &	 0.000 \\ \hline
\textbf{SEA}&	0.675 &		0.000	&	-0.011 &	 -0.001 \\ \hline
\textbf{SEC}&	0.686 &		0.000	&	-0.004 &	 -0.040 \\ \hline
\textbf{CNN MAA}&	0.680 &		-0.001	&	-0.005  & -0.076 \\ \hline
\textbf{TDNN} &	0.690 &	-0.093	&	-0.067 &  -0.056 \\ \hline
\textbf{RESNET}&	0.656 &	-0.018	&	-0.051  & -0.056 \\ \hline
\textbf{wav2vec}&	0.358 &	+0.084	&	0.000  & -0.082 \\ \hline

\end{tabular}
}

\label{table:Defenses2}
\end{minipage}
\hspace{.5cm}
\begin{minipage}[c]{.48\textwidth} 
\centering
\captionsetup{width=0.9\linewidth}
\captionof{table}{MCC and NMI for baseline classifier performance and when the classifiers were presented with EON-modified audio samples}
\resizebox{0.9\linewidth}{!}{

\begin{tabular}{|l|c|c|}
\hline
\textbf{Classifier} &		\textbf{Mean MCC - EON}&		\textbf{Mean NMI - EON}\\ \hline
Data Flair&	0.093&		0.174\\ \hline
SEA&	0.290&		0.241\\ \hline
SEC	&	0.144&	0.089\\ \hline
CNN MAA	 &	0.220 &		0.243 \\ \hline
TDNN	 &	-0.135 &		0.150 \\ \hline
RESNET	 &	-0.121 &		0.108 \\ \hline
wav2vec	 &	-0.289 &		0.361 \\ \hline

\end{tabular}
}
\label{table:MCC}
\end{minipage}

\subsubsection{Passive Defense - True Class Inference}\label{section:defenses1}

\textbf{Passive} defenses attempt to infer the true class of an audio sample without modifying the audio. The \textit{SER Provider} could exploit strong relationships between false and true classes to defeat DARE-GP. As mentioned in Section \ref{section:threats}, the evaluation metrics used to assess the relationships between the true emotion and the false labels due to DARE-GP are MCC \cite{chicco2020advantages} and NMI \cite{estevez2009normalized}. 

The MCC is a special case of the Pearson Correlation Coefficient used for classification problems. As with the Pearson Correlation Coefficient, the score is in the range [-1, 1], where 1, -1 and 0 are strong positive correlation, strong negative correlation and uncorrelated. NMI is scaled in the range [0, 1]; a value of 1 indicates that one can create a perfect one-to-one mapping from predicted class labels to actual class labels. A value of 0 indicates that there is no dependency between the predicted and actual labels. MCC and NMI values for the black-box classifiers were calculated for unperturbed audio samples and when samples were perturbed by an EON; these values are shown in Table \ref{table:MCC}.

The application of an EON significantly degraded both the MCC and NMI between true and predicted classes for all of the classifiers. These values of NMI and MCC indicate very weak correlations between the true and predicted classes. In the case of CNN MAA, over 68\% of the misclassifications were predicted as \textit{angry} or \textit{surprised}. The SEC classifier misclassified over 80\% of the EON-perturbed audio samples to \textit{sad} or \textit{angry}. The RESNET and SEC classifiers, on the other hand, strongly favored one class for misclassification (\textit{sad} and \textit{happy}, respectively). The Data Flair, SEA, TDNN SER classifiers, on the other hand, did not strongly favor any class for misclassification; these classifiers' misclassifications were spread out fairly evenly across all possible classes. Finally, in wav2vec misclassifications were split between \textit{happy}, \textit{angry}, and \textit{neutral} with rates of 52\%, 28\%, and 27\%, respectively.

Our empirical evaluations did not identify any significant inter-emotion relationship across the various classifiers used for evaluation. In addition, no clear relationship was observed between a user's true emotion and the misclassified emotion for a specific classifier. Further investigation into why these classifiers' misclassifications fell into these distributions is left as future work. The main takeaway from these results is that the EON-driven misclassifications do not reveal information that would allow the \textit{SER Provider} to reverse engineer the actual classes from these forced misclassifications.

\subsection{Acoustic EON Evaluations}\label{real-alexa-evaluation}
\begin{figure}
\begin{center}
\begin{tabular}{ccc}
\includegraphics[height=1.3in]{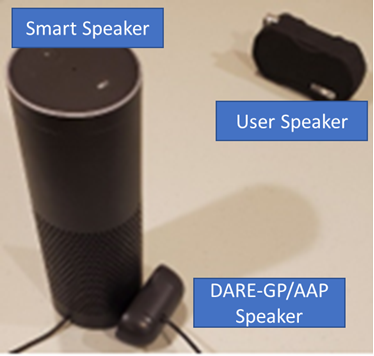} & 
\includegraphics[height=1.3in]{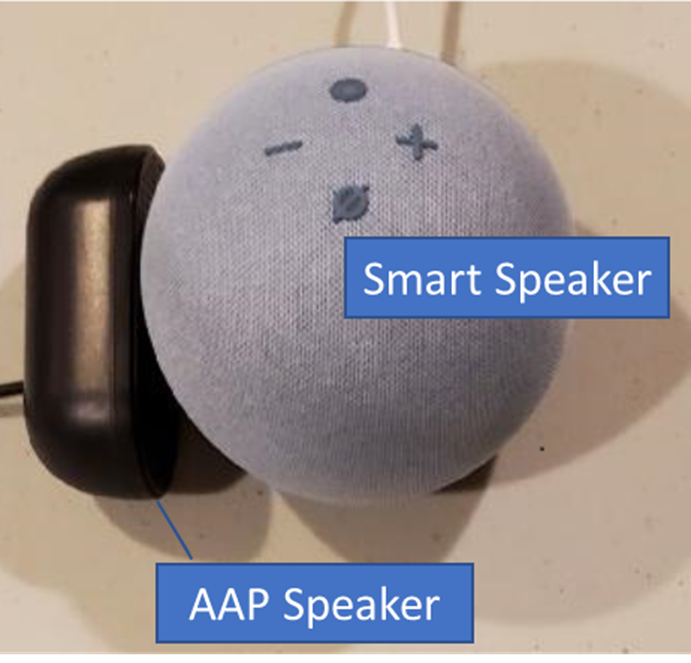} &
\includegraphics[height=1.3in]{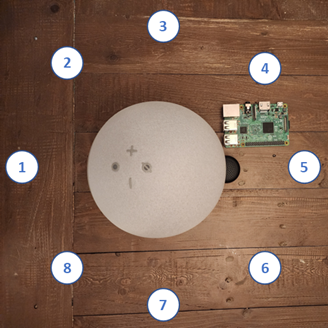} \\
\textbf{A} & \textbf{B} & \textbf{C} \\
\end{tabular}
\end{center}
\caption{ DARE-GP (A) evaluation setup with Amazon Echo. (B) evaluation setup with Amazon Dot. (C) evaluation setup with Amazon Dot and DARE-GP device implemented with a Raspberry Pi. Direction markings 1-8 used for EON direction evaluation. } 
\label{fig:speakers}
\vskip -2ex
\end{figure}

The final set of evaluations assessed whether or not a digitally-generated EON could be effective as over-air, playable sounds. ESR is the metric used for these evaluations. Since these evaluations involved commercial smart speakers, the ESR calculation used the Alexa-based speech-to-text implementation rather than the vosk/kaldi libraries used in the rest of this work. Evaluations were performed against two commercial smart speakers as shown in Figures \ref{fig:speakers} (A) and (B).

In all of these configurations, a DARE-GP/EON Speaker was placed against the smart speaker; whenever a user's speech was presented to the smart speaker, the DARE-GP/EON speaker played the EON. This configuration allowed an EON to propagate to the smart speaker via conductance of the materials in the smart speaker rather than only through the acoustics of the room, which can cause digitally-developed evasions to fail in an acoustic setting \cite{li2020practical}. The evaluations \textit{using real smart speakers} presented in this section are:

\begin{itemize}[leftmargin=0.4cm]
\item Real-time playback of EONs with user utterances from different distances (Section \ref{subsect:acousticaap})
\item Real-time playback of EONs when the user is speaking from different directions (Section \ref{subsect:directions})
\item Automated, real-time EON invocation triggered by the smart speaker's wake word (Section \ref{subsect:dist2})

\end{itemize}

\subsubsection{Acoustic Evaluation on Commercial Smart Speakers}\label{subsect:acousticaap}

\par
The first set of evaluations considers the effectiveness of a digitally-derived EON when adapted and played in an over-air acoustic setting for target users \textbf{U}. User utterances were played via a speaker at varying distances from the smart speaker (see Figure \ref{fig:speakers} (A)). These evaluations were performed using two different Alexa-based smart speakers. Recordings were captured by the smart speaker (Amazon Alexa) using a custom Alexa skill that treated each of the EON or audio sample recordings as an interaction. These audio files and corresponding Alexa transcripts were manually downloaded from the Alexa console \cite{alexaasr} (additional details provided in Appendix \ref{appendix:acousticrecordings}). The ESR was then calculated for the final EON by counting any misclassified sample as a successful evasion if the transcript extracted by Alexa was correct. 

\vskip 3ex
\begin{table}[]
  \centering 
  \caption{ESR and WER values for acoustic mixing of EONs with Amazon Echo and Amazon Echo Dot at variable distances. Train RAVDESS 40 generations, TESS 10 generations. TESS train/test/eval split of 25/25/50. Alexa transcripts were used for ESR and WER calculations.}
  \resizebox{0.99\linewidth}{!}{
  \begin{tabular}{|l|c|c|c|c|c|c|c|c|c|c|} 
    \hline  
     & \multicolumn{2}{c|}{\textbf{1ft}} 
     & \multicolumn{2}{c|}{\textbf{2ft}} 
     & \multicolumn{2}{c|}{\textbf{3ft}} 
     & \multicolumn{2}{c|}{\textbf{5ft}} 
     & \multicolumn{2}{c|}{\textbf{9ft}}  \\
    & ESR & WER & ESR & WER & ESR & WER & ESR & WER & ESR & WER \\
    \hline
    Amazon Echo 
    & $0.815 \pm 0.014$ & 0.012
    & $0.82 \pm 0.018$ & 0.013
    & $0.756 \pm 0.022$ & 0.011
    & $0.743 \pm 0.02$ & 0.0187
    & -- & -- \\
    \hline  
    Amazon Echo Dot 
    & $0.812 \pm 0.03$ & 0.012
    & $0.804 \pm 0.031$ & 0.012
    & $0.769 \pm 0.011$ & 0.016
    & $0.738 \pm 0.04$ & 0.015
    & $0.71 \pm 0.053$ & 0.013 \\
    \hline  
  \end{tabular}
  }
  \label{table:eonmultidist}
\end{table}

Table \ref{table:eonmultidist} shows the evaluation results for the configurations shown in  Figure \ref{fig:speakers} (A) and (B) when the "simulated-user (i.e., User Speaker)" was 1ft away from the smart speaker. Both had comparable SER classifier evasion rates (around 0.9), but with low word error rate (WER; a measure of transcription accuracy), likely due to the robust microphone arrays present in smart speakers. This low WER indicates that the EONs are not overpowering the user utterances. Since these were the first trials to play EONs acoustically, this was the first opportunity to assess how invasive the EONs were. During these evaluations, the mean background noise in the room was 46dB. The loudest EON generated in these experiments was 50dB which is below the normal conversation range \cite{humanDB}. 

These evaluations were repeated with the "simulated-user" (i.e., User Speaker) at different distances. The EONs and loudness were kept constant at each distance to assess how well these EONs would work if the user were interacting with the smart speaker from different locations in the room. As shown in Table \ref{table:eonmultidist}, even at a distance of 9ft, there is minimal degradation in the EON's effectiveness. The reason for such robustness is the DARE-GP’s position with respect to the smart speaker. Irrespective of ``user's'' distance from the speaker, the EONs are always played from a device touching the smart speaker; this means that the smart speaker only needs to be robust with respect to capturing the user's speech from different ranges, which is a primary requirement of these devices.  

\subsubsection{Evaluation of EON Robustness from Different Directions}\label{subsect:directions}

In addition to distance, the direction from which a user speaks could also impact the effectiveness of an EON due to considerations like beamforming. To evaluate this, the same audio samples used in Section \ref{subsect:acousticaap} were played at 5ft and 9ft from the smart speaker at each of the directions shown in Figure \ref{fig:speakers} (C). The EON was played from the small speaker adjacent to the smart speaker as shown in Figure \ref{fig:speakers} (C), with no change to EON or utterance loudness at the respective speakers. The resulting ESR of this EON at each distance/direction combination is provided in Table \ref{table:directions}.

\begin{table}[]
  \centering 
  \caption{ESR values for an EON where the direction of the user with respect to the smart speaker and the speaker playing the EON changed. See Figure \ref{fig:speakers} (C) for details.}
  \resizebox{0.6\linewidth}{!}{
  \begin{tabular}{|c|c|c|c|c|c|c|c|c|c|c|} 
    \hline
    & \multicolumn{8}{c|}{\bfseries Directions} & \bfseries Mean & \bfseries Variance\\
     & \bfseries 1 & \bfseries 2 & \bfseries 3 & \bfseries 4 & \bfseries 5 & \bfseries 6 & \bfseries 7 & \bfseries 8 &  &  \\
    \hline
    5ft & 0.876 &  0.852 & 0.848 & 0.864 & 0.86 &  \textbf{0.836} & 0.844 & 0.852 & 0.854 & 0.00012 \\
    \hline  
    9ft & 0.72 &  0.708 & 0.74 &  0.764 & \textbf{0.604} & 0.656 & 0.772 & 0.752 & 0.7145 & 0.00294 \\
    \hline  
  \end{tabular}
  }
  \label{table:directions}
\end{table}

User direction with respect to the smart speaker and the DARE-GP device does play some role in EON effectiveness. The trials in Table \ref{table:directions} showed a slight degradation when the user was aligned closely with the DARE-GP speaker. This added relatively higher disruption on the speech signal, causing higher degradation on transcription efficacy, resulting in lower ESR. The effect was much more pronounced in the 9ft test, which was not surprising considering the larger variance observed at 9ft in the distance evaluation (see Table \ref{table:eonmultidist}). That said, this trial demonstrates that EONs are robust to user direction even when the smart speaker uses beamforming (the Echo Dot) to focus the microphone array on the direction from which the user is speaking. We believe this robustness is due to the DARE-GP speaker's physical contact with the smart speaker, enabling propagation of EON via conductance of the materials in the smart speaker. EON is not being presented to the smart speaker microphone over the air only from one direction, debilitating any directional noise masking by the beamforming.

\subsubsection{Automated EON Invocation}
\label{subsect:dist2}

The previous evaluations demonstrated that DARE-GP is effective in an acoustic setting. To deploy a pre-trained DARE-GP system (one in which an EON has already been developed) three components are required: a small processor, a microphone, and a speaker. The processor is responsible for listening (i.e., detecting) for the smart speaker's wake word via the microphone and then playing the pre-calculated EON at a pre-determined volume through the speaker. This deployment configuration should introduce a small amount of lag between when the user starts to speak and when the EON starts to play; this evaluation measured the impact of this lag on EON effectiveness. 

To demonstrate this setup, the final evaluation used a Raspberry Pi 3, Model B with a Quad Core 1.2GHz Broadcom BCM2837 64bit CPU and 1GB of RAM, along with an off-the-shelf speaker and microphone (see Figure \ref{fig:speakers} (D)). An off-the-shelf wake word detection engine \cite{porcupine} was used to listen for "Alexa" before playing the EON. In this configuration, DARE-GP consumes less than 3.7 W of power during peak usage. The EON used for this evaluation used the same EON and volume as the first trial at 1ft using the Amazon Dot from Section \ref{subsect:acousticaap}.

The original \emph{ESR} for this trial was $0.76$. When executing the same EON in a fully automated manner, the resulting \emph{ESR} was $0.738$. The average time between wake word detection and the start of EON playback was $0.15$ seconds. This evaluation, along with the previous ones, demonstrate that DARE-GP's digitally-generated EONs can transition successfully to an acoustic setting with multiple smart speakers in a realistic setting. In fact, this approach benefits from characteristics of the realistic setting (i.e. - the superior microphones and speech-to-text utilities in the smart speakers), resulting in a very effective solution. Further, it demonstrates that this approach's use of a pre-generated, universal EON allows real-time deployment on a very small form factor.

\section{Discussion}\label{section:discussion}

There were some important experiment design choices and constraints that are worth discussing in more detail:

\textbf{Use of Recorded Sound and Off-the-Shelf SER Classifiers}: For this work, evaluations used previously-recorded audio. This approach provided a benchmark against a set of known datasets, rather than possibly introducing some inadvertent bias that would lead to unrealistic results. Likewise, the off-the-shelf SER classifiers, were used for a similar reason. The use of existing data and existing SER classifiers demonstrates the ability of DARE-GP to generalize. 

\textbf{Surrogate SER Classifier}: The surrogate SER classifier used to train DARE-GP is significant because of its distinction from the black-box classifiers. This classifier is a simple DNN that uses MFCC features. While the .877 accuracy is acceptable for learning the emotion-relevant spectral speech attributes, it is not as high as some of the black-box classifiers. Additionally, it has different classes than Flair and SEA, and different features than Flair, SEC, RESNET and TDNN. This is important concerning the utility of DARE-GP; by targeting an underlying attribute of a user's speech, the spectral components, DARE-GP was effective despite the differences between the surrogate and black-box SER classifiers' implementation and performance. 

\textbf{Targeted versus Untargeted Evasion}: DARE-GP is an example of \textit{untargeted evasion}; an evasive sample has fooled the SER classifier as long as the predicted class is different from the actual class. Based on the objectives of DARE-GP, untargeted evasion was sufficient. As demonstrated in Section \ref{section:defenses1}, the actual and false classes are sufficiently independent of one another, which would prevent the \textit{SER Provider} from inferring the actual emotion associated with a given utterance. This means, for example, that no successful targeted advertising is possible utilizing the emotional content of any speaker's utterance. 

\begin{figure}
\begin{center}
\begin{tabular}{cc}
\includegraphics[height=1.3in,trim={1cm 2cm 6cm 11cm},clip]
{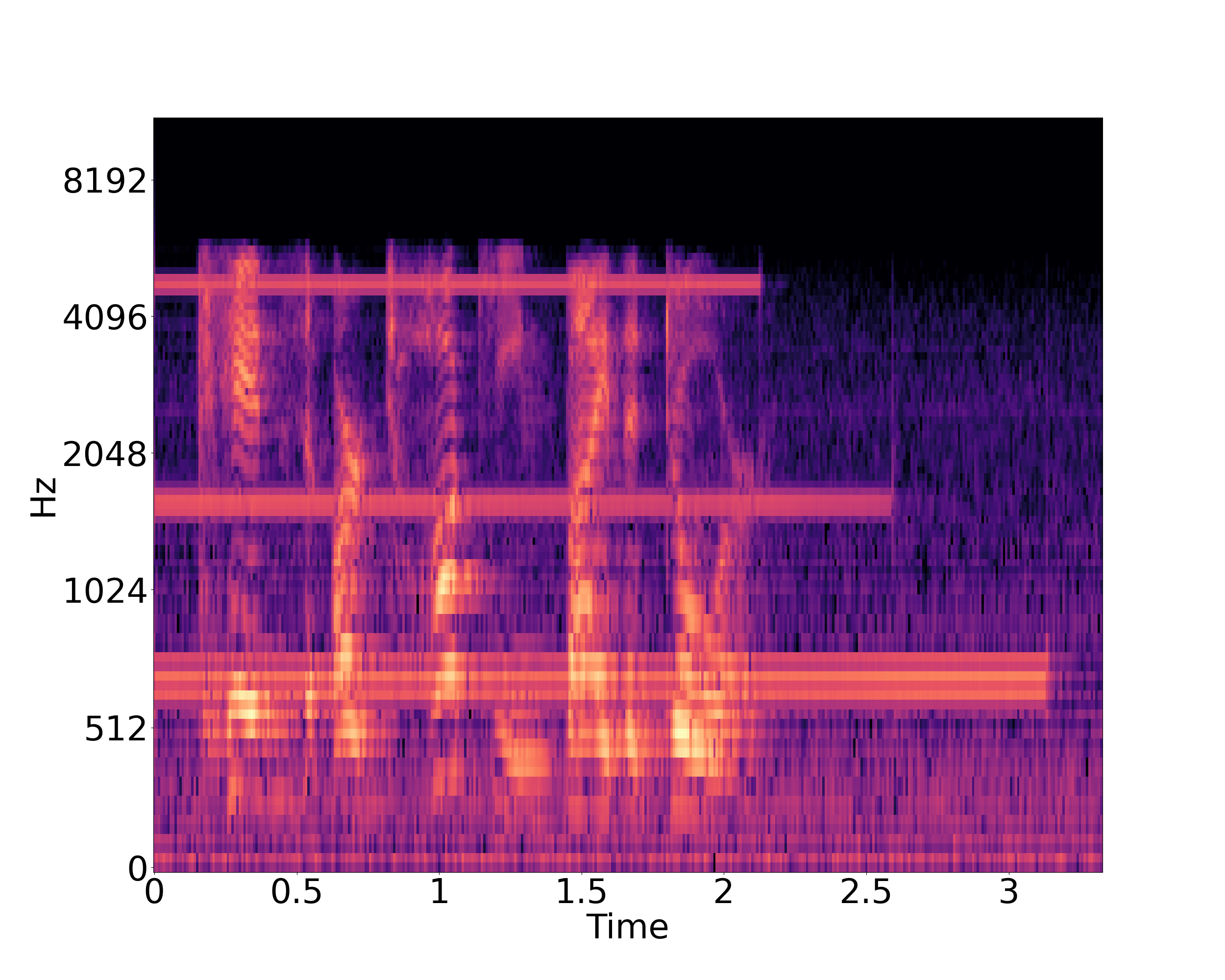} & 
\includegraphics[height=1.3in,trim={1cm 2cm 6cm 11cm},clip]
{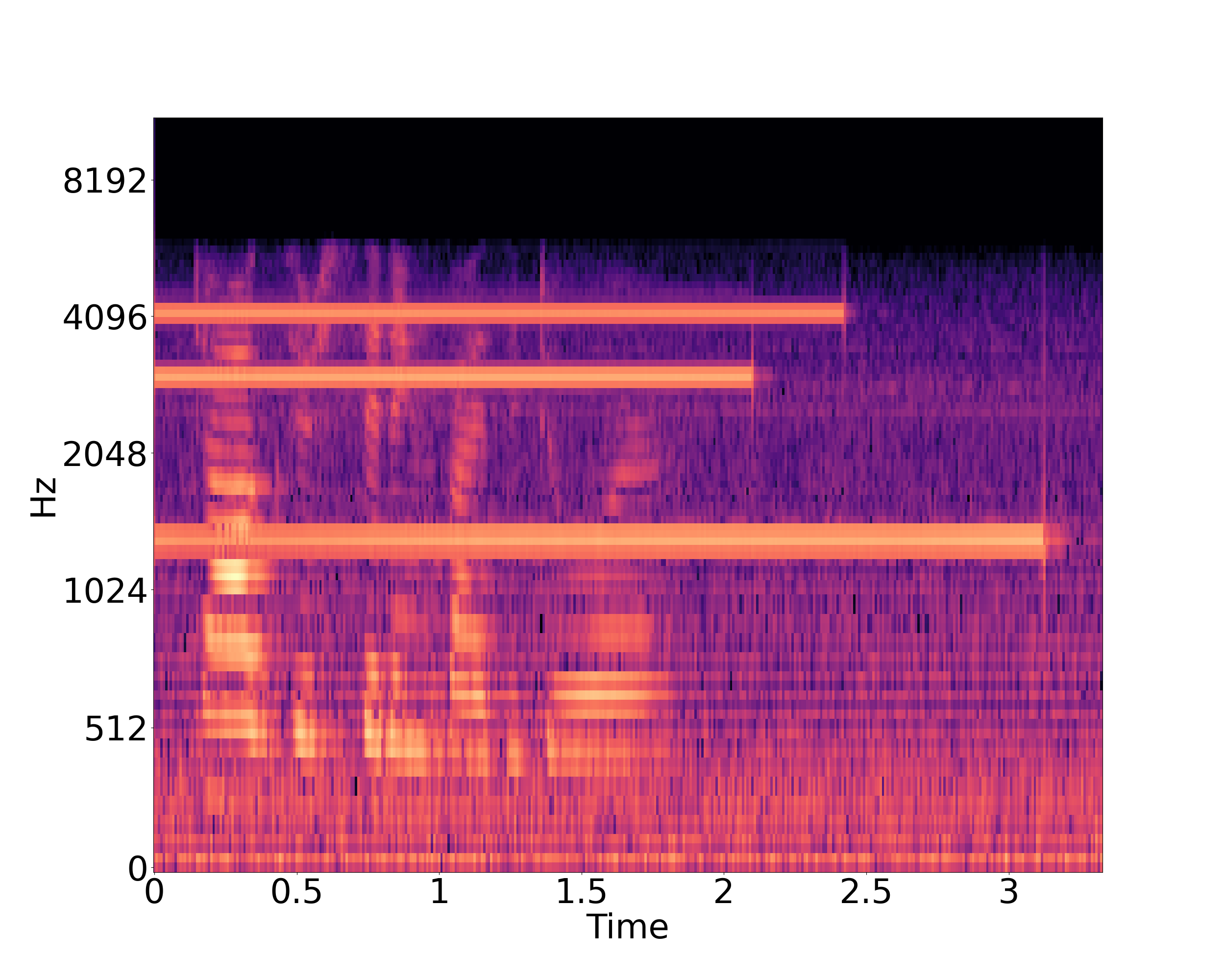}\\
\textbf{A} & \textbf{B} \\
\end{tabular}
\end{center}
\vskip -2ex
\caption{ Spectrograms of sample user utterances mixed with EONs generated for (A) a group of three men and two women, and (B) a group of five women. } 
\label{fig:spectrograms}
\end{figure}

\textbf{Differences Across EONs}: The final EON for a given set of users is tailored based on the specific users’ common emotion-relevant spectral traits. Figure \ref{fig:spectrograms} shows two spectrograms trained for different sets of users, varied in demographics from the RAVDESS dataset. Figure \ref{fig:spectrograms} (A) shows an EON that overlaps formant $F3$ and $F6$ and one tone in a higher register. In Figure \ref{fig:spectrograms} (B), the EON masks formant $F4$ and also masks two higher frequency bands. This figure provides an example that DARE-GP learns different EONs for different sets of users. In this example, two groups (i.e., sets of users) with a demographic difference may have differences in their common emotion-depiction-relevant spectral traits. Hence the generated EONs are different. 

Notably, initial attempts to generate EONs that would work for previously unseen users were unsuccessful; cross-dataset evaluations where RAVDESS-trained EONs were applied to TESS audio samples showed that some tailoring was required. While this approach \textit{does} require labeled audio samples for the end users, using a pre-trained model to label unlabeled user data, along with approaches to measure the adaptation performance of such an untrained model \cite{gong2023dapper} could decrease the impact of this issue. Such investigation is left as future work.

\textbf{User Survey}: In addition to the evaluations performed in Section \ref{section:evaluation}, a survey was performed with $19$ undergraduate and graduate students. This survey aimed to 1) assess interest in emotion protection, 2) get user feedback on the user-friendliness of DARE-GP, and 3) the user-friendliness of a notional system that implements a record$\rightarrow$perturb$\rightarrow$replay design. When asked about the level of concern over companies using their emotion data for targeted ads, the mean level of concern was $3.16$ out of a maximum of $4.0$, with only two respondents rating their concern as \textit{Neutral} or \textit{Unconcerned}. Only $1$ out of the $19$ respondents rated the EON used for evaluation in Section \ref{subsect:dist2} as  \textit{Somewhat Distracting}. Most of the SotA approaches discussed in Section \ref{section:related} would require recording a user's utterance, modification of the utterance to generate an evasive variant, and then playback to the smart speaker. To simulate this, a script was used to record the user's utterance after hearing a wake word. Then, the sample was modified slightly, and then played after playing a pre-recorded wake word for the smart speaker. $36.8\%$ of the respondents stated that they would be unwilling to use this system. This survey provided further evidence that DARE-GP is both relevant and more usable than SotA systems. Additional details are provided in Appendix \ref{appendix:survey}.

\textbf{Beamforming Countermeasure}: The current approach addresses beamforming via physical contact between the DARE-GP speaker and the smart speaker. Future smart speakers, if manufactured to limit the impact of external vibrations, could potentially improve the effectiveness of beamforming by limiting EON presentation via conductance. Since no commercial smart speakers currently advertise this feature, evaluating the effectiveness of such a defense is beyond the scope of this paper and is left as future work.

\textbf{Extended, In-the-wild Evaluation:} Evaluations were performed in laboratory environments and in two home settings; exhaustive in-the-wild evaluation was out of scope. Future work would benefit from sampling data from a broader range of in-the-wild situations to determine DARE-GPs effectiveness and any long-term EON invasiveness concerns.

\section{Related Works}\label{section:related}

There are three classes of related works relevant to DARE-GP: (1) works that protect against unanticipated recording, (2) GP-based ML evasion, and (3) audio evasion works. 

\subsection{Protection of Audio against Unanticipated Recording}\label{section:protectvsrecording}

There are several works that address protection from unanticipated recording. These works include both protection from unauthorized devices and cases where the user wants to prevent known devices from unauthorized recording. Patronus \cite{li2020patronus} is an approach to perform real-time audio jamming that could only be decoded by devices with the correct "key." MicShield \cite{sun2020alexa} prevents smart speakers from recording audio outside the window of a command being directed at the smart speaker. In both cases, the solutions presented prevent access to any audio by unanticipated devices at unplanned times. DARE-GP is complementary to these approaches by providing protection with respect to emotional content for devices intended to hear a given command.   

\subsection{Evasion of Black-Box Classifiers Using Genetic Programming}\label{section:advevasion}

Works related to evading classifiers without requiring detailed insight into the inner-workings of the evaded classifiers are generally referred to as black-box adversarial evasion \cite{tabassi2019taxonomy}. The use of genetic programming (GP) in DARE-GP was inspired by previous applications of GP towards black-box adversarial evasion in multiple domains, including: PDF malware \cite{weilin2016automatically}, WiFi behavior detection \cite{liu2022physical}, and audio speech recognition (ASR) \cite{taori2019targeted, alzantot2018did}. The primary difference between DARE-GP and these existing works is that DARE-GP uses GP to generate \textbf{universal} adversarial perturbations (UAPs); that is, the EON generated by DARE-GP causes misclassification in previously unseen inputs (utterances). All of these other works used GP to generate unique perturbations for each evasive sample. Unlike these existig approaches, by precomputing a single, universal adversarial perturbation, DARE-GP is able to operate \textit{in real-time} for \textit{previously unheard utterances}.
Additionally, compared to existing GP-based audio evasion approaches \cite{wu2019semi,alzantot2018did}, DARE-GP leveraged constrained GP to protect the speech-to-text transcription utility, making it practically deployable for the target task against commercial smart speakers.

\subsection{Adversarial Evasion of Audio Classifiers}\label{section:advevasion}

\begin{table}
  \caption{Comparison of related works.}
\resizebox{0.8\linewidth}{!}{
\begin{tabular}{|l|c|c|c|c|c|c|}
\hline
&	\textbf{Transferable to} &	
\textbf{Protects}&	\textbf{Effective} &	
\textbf{Real-Time for Previously} &	
\textbf{Natural Interaction} &
\textbf{Robust vs.} \\
& \textbf{Black Box Model} & \textbf{Transcription}& \textbf{Acoustically} & \textbf{Unheard Utterances} &\textbf{with Smart Speaker}& \textbf{Defenses}
\\ \hline

DARE-GP &	$\checkmark$ &	$\checkmark$ &	$\checkmark$  &	$\checkmark$ & $\checkmark$ & $\checkmark$ \\ \hline

Advpulse \cite{li2020advpulse}&	  &	 &	$\checkmark$  &	$\checkmark$ & $\checkmark$ & $\checkmark$ \\ \hline

EDGY \cite{aloufi2020paralinguistic}& $\checkmark$  & Partial &	$\checkmark$  &	$\checkmark$ &  &  \\ \hline

Spectral Envelope \cite{gao2022black}& $\checkmark$  &  &	  &	 &  &  \\ \hline

SMACK \cite{yu2023smack} &   &  & $\checkmark$  &	 &  & $\checkmark$ \\ \hline

Practical (RIR) \cite{li2020practical}&   & Partial & $\checkmark$  &	 &  &  \\ \hline

Imperio \cite{schonherr2020imperio}&   &  & $\checkmark$  &	 &  &  \\ \hline

Semi-Black-Box \cite{wu2019semi}&   & Partial &  &	 &  &  \\ \hline

Targeted Black-Box \cite{taori2019targeted}&   &  &  &	 &  &  \\ \hline

Did You Hear That? \cite{alzantot2018did}&   &  &  &	 &  &  \\ \hline

\end{tabular}
}
  \label{table:relatedworks}
\end{table}

There are many works that consider evasion of audio classifiers, including speaker recognition \cite{abdoli2019universal, kajarekar2006study, li2020advpulse, aloufi2020paralinguistic}, automated speech recognition (ASR) \cite{abdullah2021sok, taori2019targeted, abdoli2019universal, carlini2018audio, li2020advpulse, alzantot2018did, schonherr2020imperio, wu2019semi, yu2023smack}, and speech emotion recognition (SER) \cite{aloufi2019emotionless, gao2022black, aloufi2020paralinguistic}. Since nothing in these approaches intrinsically precludes application to other audio domains, these were all considered when comparing DARE-GP to SotA techniques. However, as shown in Table \ref{table:relatedworks}, there are several key attributes that differentiate these works from DARE-GP. 
 
Advpulse \cite{li2020advpulse} presented an approach for targeted evasion of speaker recognition and automated speech recognition classifiers. Of the considered works, Advpulse has multiple similarities with DARE-GP. Advpulse generates additive noise in the form of $0.5$ second pulses; these pulses are universal adversarial audio perturbations, much like DARE-GP's EONs. As a result, like DARE-GP, Advpulse can be used for real-time evasion of previously unheard utterances in an acoustic setting. The authors were concerned that the additive noise might only be effective if played at exactly the right time to cause misclassification with an utterance. To address this, the authors built in variable delays in their training data to address this concern. DARE-GP did not address this explicitly; mainly because the use case presented provides a simple way to align EONs with user speech: the wake word for the smart speaker's VA. Details on how this worked are provided in Section \ref{subsect:dist2}. Advpulse also considered the possibility of a knowledgeable adversary attempting to thwart their evasion approach; the defenses presented are very similar to those presented in Section \ref{section:defenses}. There are two major differences between DARE-GP and Advpulse: \textit{transcription protection} and \textit{black-box transferability}. Advpulse did not consider inadvertent impacts on speech transcription from their approach. More importantly, Advpulse is a white-box evasion system; the approach required detailed access to the target model's gradients and weights in order to develop their additive noise. Both of these limitations of this work make Advpulse unsuitable for the use case presented in this paper. 

EDGY \cite{aloufi2020paralinguistic} considers the privacy of a user's speech as a challenge of separating the text of what a user says (linguistic) from the other sensitive user information carried in speech like the user's gender or emotion (paralinguistic). To address this, the authors learned separate, disentangled representations (embeddings) of the linguistic and paralinguistic parts of a user's speech. A significant part of this work was devoted to how to perform this separation at edge devices in a resource-constrained environment. One major benefit of this work is that it does not require access to a target classifier to develop these embeddings. The authors developed several classifiers for paralinguistic information based upon a TRIpLet Loss network (TRILL) \cite{shor2020towards} and demonstrated that using the linguistic embedding could significantly degrade the classification of multiple private user attributes, including age, emotion, accent, and gender. The most significant difference between this work and DARE-GP is the employment scenario. DARE-GP EONs are additive noise that can be played in real-time when a user speaks. EDGY, like many adversarial audio works \cite{gao2022black, yu2023smack, li2020practical, schonherr2020imperio, wu2019semi}: 1) requires the user to speak, 2) processes the audio to generate the adversarial sample, and 3) uses a vocoder or similar capability to generate synthetic speech to evade the classifier. This impacts both usability of the system and can degrade audio quality, as shown in the impacts on word error rate (WER) for the generated audio. 

Gao et al. \cite{gao2022black} demonstrated that three spectral envelope-based attacks could defeat several previously unseen SER classifiers. Like DARE-GP, these attacks were computed without any insight into the target classifiers. These attacks were
not executable in real-time; unlike DARE-GP’s EONs, the spectral envelope filters applied (vocal tract length normalization, modulation spectrum smoothing, and McAdams transformation) cannot be represented as additive noise that can be played when a user speaks. To deploy these filters in a real-world scenario, the users’ utterances would need to be recorded, modified, and then replayed within range of the target smart speaker. Further, these
filters introduced significant transcription errors, measured as Word Error Rate (WER). Evaluations comparing DARE-GP's performance to these methods are presented in Section \ref{section:evalcompare}.

SMACK \cite{yu2023smack} performed evasion against several automated speech recognition (ASR) and speaker recognition (SR) classifiers by modifing the prosody of speech to disrupt these classifiers while attempting to maintain "natural sounding" speech. Like several other state-of-the-art (SotA) works \cite{li2020practical, yu2023smack, schonherr2020imperio, li2020advpulse}, SMACK employed gradient-based methods in feature space to generate adversarial samples. Like multiple SotA works, SMACK leveraged room impulse response (RIR) \cite{stan2002comparison} to model a room's acoustics in order to improve the success rate of these digitally-created evasive samples when playing them in acoustically. Like \cite{gao2022black} and \cite{aloufi2020paralinguistic}, SMACK is not real-time and would require preprocesing of a user's audio before presenting it to the smart speaker's VA. 

The remaining works are similar to those presented above. Imperio \cite{schonherr2020imperio} evades a white-box ASR classifier, also using RIR as an approach to improve effectiveness of evasive samples when played acoustically. Wu, Yi, et al. \cite{wu2019semi} also evaded and ASR classifier, but was able to do so without requiring the gradients and weights of the target model. This work used GP to generate unique adversarial perturbations for each sample, but could not operate in real-time and did not evaluate the efficacy of these samples when played acoustically. Similarly, Alzantot et al. \cite{alzantot2018did} also evaded ASR systems using GP, but with black-box access to the target model (hard label access). Finally, Taori et al. \cite{taori2019targeted} also used  GP to perform targeted evasion against an ASR system. This work used a new mutation operation based on gradient estimation to speed convergence. Like other SotA GP-based black-box evasion works, this work required hard-label access to the target classifier and generated unique perturbations for each utterance in a non-real-time scenario.

Each of these existing works manifests some of the characteristics of a usable system to address the challenges laid out in this paper. That said, each of them falls short in one or more area that would make the system unusable for the use case described in this work.

\section{Conclusion}\label{section:conclusions}
The objective of this work was to degrade the privacy loss concerning speaker emotions incurred by using a smart speaker VA service (by an \emph{SER Provider}). The paper presents DARE-GP, which for a set of users, precomputes an EON (i.e., universal audio perturbation) that disrupts emotion detection by adding spectral distortions to the target users’ speech, hence effective against previously unheard utterances and transferable to broad set of black-box SER classifiers with heterogeneous classes, topologies, and feature representations. Generated EON is additive, meaning that simultaneously playing an EON with the user's speech can evade SER classifiers over the air in real-time. Our extensive evaluation shows DARE-GP's superior performance compared to state-of-the-art SER attacks and robustness against defenses from a knowledgeable \textit{SER Provider}. Finally, DARE-GP was demonstrated effective in a real-time, real-world scenario against commercial smart speakers, where a SWaP-efficient form factor using wake word detection could automatically deploy an EON whenever a user interacted with a smart speaker, demonstrating an ESR of 0.738 against an Amazon Dot. 

\section*{Acknowledgments}\label{section:acknowledgements}
This work was supported in part by NSF IIS SCH \#2124285. In addition, we extend our gratitude to the Air Force Research Laboratory Information Directorate for supporting this work.

\bibliographystyle{ACM-Reference-Format}
\bibliography{main}

\appendix

\section{Sentiment Analysis Results}\label{appendix:sentiment}

In generic human conversations, emotional information can be leaked via an utterance's transcript \cite{alswaidan2020survey}. DARE-GP addresses leakage of emotion information from the spectral features of speech, but does not consider the loss of emotional privacy based upon the text content extracted from smart speaker commands. To evaluate if the text content was a significant source of accurate emotional information, a set of $88$ common Alexa commands \cite{alexacmds}, both phrases from the Ryerson Audio-Visual Database of Emotional Speech and Song (RAVDESS) dataset \cite{livingstone2018ryerson}, and $20$ utterances from the Toronto Emotional Speech Set (TESS) \cite{dupuis2010toronto} were presented to two different sentiment analysis classifiers: Google Cloud Sentiment Analysis \cite{googlesentiment} and DEEP API Sentiment Analysis \cite{deepapi}. 

DEEP API labeled $50 of 106$ as \textit{Not Neutral}. Some representative examples include:

\begin{itemize}
    \item Alexa, unmute. ['Positive']
    \item Alexa, stop ['Negative']
    \item Alexa, shut up. ['Positive']
    \item Alexa, show me the trailer for It. ['Negative']
    \item Alexa, play Pitbull's on Spotify. ['Negative']
\end{itemize}

Google's Cloud Sentiment Analysis scores in a range of [-1.0, 1.0] between very negative and very positive. Out of 106 utterances, 8 not neutral with confidence > 0.5:

\begin{itemize}
    \item say the word burn, Negative -0.7
    \item Alexa, stop Negative -0.6
    \item Alexa, help me wash my hands. Positive 0.6
    \item Alexa, cancel the pizza timer Negative -0.6
    \item Alexa, cancel my alarm for 2 p.m. Negative -0.7
    \item Alexa, buy this song Positive 0.7 
    \item Alexa, buy this album. Positive 0.6
    \item Alexa, find me a good smartphone on Amazon, Positive 0.7
\end{itemize}

It is not entirely clear why some of these commands were listed as positive or negative. The TESS utterances, for example, are intentionally neutral to prevent inference of emotion from the transcripts, yet "say the word burn" was considered negative by Google's Cloud Sentiment Analysis. The bottom line is that these sentiment analysis inferences were at best inconsistent when working on such short transcripts. 

\section{Speech-Emotion-Recognition (SER) Classifier Details}\label{appendix:classifierdetails}

This work includes evaluations against a broad set of SER classifiers with heterogeneous features, topologies, and classes. Additional details on these classifiers are provided below.

\begin{table*}
\caption{Details regarding black-box SER classifiers.}
\resizebox{0.9\linewidth}{!}{
\begin{tabular}{|p{5.5cm}|p{8.4cm}|p{4cm}|p{1.3cm}|}
\hline
\textbf{Classifier}& \textbf{Description} & \textbf{Classes} & \textbf{Accuracy}\\ \hline
Simple DNN (the surrogate classifier)  &	A 7-layer, densely connected DNN constructed for use in the EON fitness function. Uses MFCC  features.&	8: neutral, calm, happy, sad, angry, fearful, disgust, surprised & .877 \\ \hline

Data Flair \cite{flair2019}	&2-Layer, multi-Layer perceptron classifier published as part of the data-flair machine learning training program. Uses MFCC and Chroma features. &	4: calm, happy, fearful, disgust & .665 \\ \hline

Speech Emotion Analyzer (SEA) \cite{sea2021} &	An 18-Layer, convolutional neural network (CNN) that classifies based upon gender and emotion. &	10: angry, calm, fearful, happy, sad for both male and female & .921 \\ \hline

Speech Emotion Classification - CNN-LSTM (SEC) \cite{kosta2021} &	Bi-directional LSTM attention mechanism and a CNN using Mel Spectrogram features. &	8: neutral, calm, happy, sad, angry, fearful, disgust, surprised & .916 \\ \hline

Conv. Neural Network with Multi-scale Area Attention (CNN-MAA) \cite{gao2022black} &	Based upon implementation by Gao et al. \cite{gao2022black} as a target for adversarial evasion, based upon Xu et al. \cite{xu2021speech}. Uses MFCC features. &	8: neutral, calm, happy, sad, angry, fearful, disgust, surprised & 0.952 \\ \hline

Time delay Neural Network (TDNN) &	A Time Delay Neural Network (TDNN) SER classifier built specifically for this work. The input to this classifier is a 3s spectrogram. &	8: neutral, calm, happy, sad, angry, fearful, disgust, surprised & 0.678 \\ \hline

Residual Neural Network (RESNET) &	A RESNET SER classifier built specifically for this work. The input to this classifier is a 3s spectrogram.  &	8: neutral, calm, happy, sad, angry, fearful, disgust, surprised & 0.773 \\ \hline

wav2vec &	A wav2vec SER classifier built specifically for this work. The input to this classifier is a 3s spectrogram.  &	4: neutral,  happy, sad, angry & 0.963 \\ \hline

\end{tabular}
}
\end{table*}

\section{Details Regarding Acoustic Evaluation Recordings}
\label{appendix:acousticrecordings}

During the acoustic evaluations performed in Section \ref{real-alexa-evaluation}, there were two types of recording actions performed: recordings for fine-tuning DARE-GP, and recordings for evaluation of DARE-GP. 

\begin{figure}
\begin{center}
\begin{tabular}{cc}
\includegraphics[height=2in]{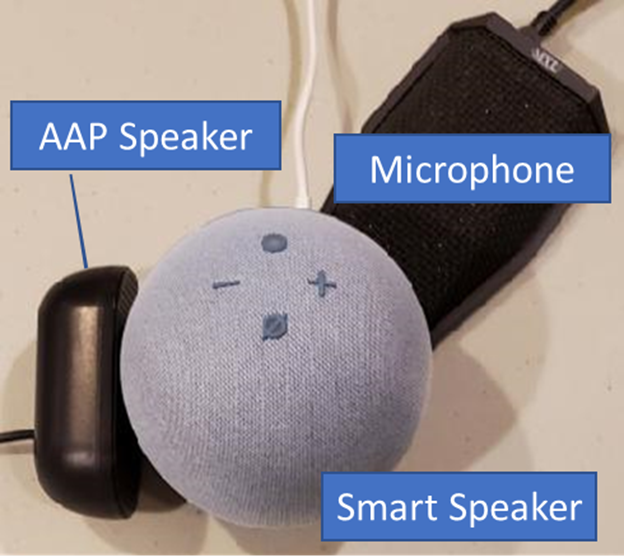} & \includegraphics[height=2in]{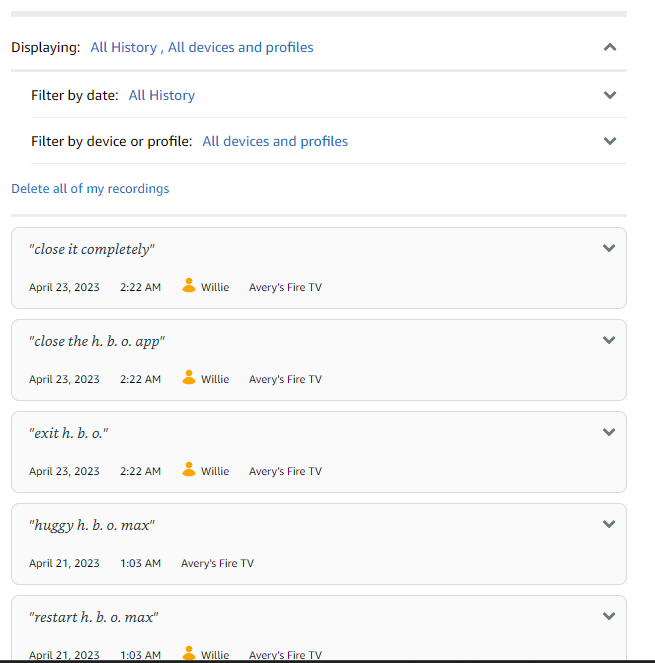} \\
\textbf{A} & \textbf{B} \\
\end{tabular}
\end{center}
\caption{ (A) Test setup with Amazon Dot and 3rd-party microphone for EON and utterance capture. This configuration was used exclusively for collection of data for EON fine-tuning and was not used for evaluation. (B) Alexa Review Voice History page. For evaluation, a script was used to retrieve the audio recordings from this page so that evaluations would be performed using Alexa-captured audio. } 
\label{fig:appendixb}
\vskip -1ex
\end{figure}

As mentioned previously, DARE-GP was pre-trained using a "canonical" dataset (RAVDESS) before fine-tuning the resulting EONs with a target household's user data (TESS). The purpose for this was to limit the amount of time and sample data required from the household when setting up DARE-GP. It took approximately 20 hours to perform 40 generations of training on RAVDESS; by fine-tuning the EONs for a mere 10 generations using user data, we were able to get similar performance without requiring extensive in-home training time. That said, in order to simulate how users in the home would provide training data for this fine-tuning process, the users would need to be able to record audio samples. The ideal approach would have been to collect these recordings using the smart speaker, but there is no Alexa API to perform this task. Rather than forcing the end users to perform a manually intensive task to collect recorded audio samples from the Alexa Review Voice History page (see Figure \ref{fig:appendixb}B), we used an off-the-shelf microphone (an MXI PoConSeries 1 AC-404USB) to capture these training audio samples. The configuration for this setup is shown in Figure \ref{fig:appendixb}A. This microphone was touching the smart speaker at a 100-degree angle (empirically determined) with respect to the speaker playing the EONs to mimic the propagation of sound through the smart speaker's materials. This setup allowed the microphone to record sound based on the room’s acoustics and the smart speaker's audio conduction \cite{weik2012communications}  without requiring any privileged access to the smart speaker. 

For evaluation, on the other hand, EON-modified recordings were harvested from Alexa. To do this, each EON-modified audio was presented to that smart speaker. Then, using a python script that leveraged both selenium and pyautogui, we downloaded the recordings from the Alexa Review Voice History page, correlated them with the originally played audio, and then performed evaluation.

\section{Usability Survey}
\label{appendix:survey}

To assess the impetus for, and the utility of, DARE-GP, we conducted a survey of $19$ college students. The questions and methodology used are presented below.

\begin{enumerate}

    \item \textbf{Context:} The respondent is informed that they will be presented with two exemplar systems intended to protect emotion information in your speech when using a smart speaker. 

\vspace{.5cm}

    \item \textbf{Question 1:} How concerned are you about companies using data about your emotions to target ads?

\end{enumerate}

\begin{tabular}{|c|c|c|c|c|}
\hline
$0$&$1$&$2$&$3$&$4$ \\
Very Unconcerned&Somewhat Unconcerned&Neutral&Somewhat Concerned&Very Concerned\\
\hline
\end{tabular}

\vspace{.5cm}

\begin{enumerate}
\setcounter{enumi}{2}

    \item \textbf{Action:} Prompt the user to say the DARE-GP wake word and a smart speaker command. This will start playing an EON as the user speaks the command. 

    \item \textbf{Action:} Prompt the user to say the "other" system wake word and a smart speaker command. This will record the user's speech, digitally modify the audio, and play back the user's command with a smart speaker wake word. 

\vspace{.5cm}

    \item \textbf{Question 2:} How distracting was the noise played by the first system?
\end{enumerate}

\begin{tabular}{|c|c|c|c|c|}
\hline
$-2$&$-1$&$0$&$+1$&$+2$ \\
Very Distracting&Somewhat Distracting&Neutral&Noninvasive&Completely Inaudible\\
\hline
\end{tabular}

\vspace{.5cm}

\begin{enumerate}
\setcounter{enumi}{5}

    \item \textbf{Question 3:} How willing would you be to use the first system?
\end{enumerate}

\begin{tabular}{|c|c|c|c|c|}
\hline
$-2$&$-1$&$0$&$+1$&$+2$ \\
Very Unwilling&Somewhat Unwilling&Neutral&Somewhat Willing&Very Willing\\
\hline
\end{tabular}

\vspace{.5cm}

\begin{enumerate}
\setcounter{enumi}{6}

    \item \textbf{Question 4:} How willing would you be to use the second system?
\end{enumerate}

\begin{tabular}{|c|c|c|c|c|}
\hline
$-2$&$-1$&$0$&$+1$&$+2$ \\
Very Unwilling&Somewhat Unwilling&Neutral&Somewhat Willing&Very Willing\\
\hline
\end{tabular}

\vspace{.5cm}

The respondents' answers to these questions are summarized in Table \ref{table:surveyresponses}. There are a few key takeaways from the survey responses. First, nearly all respondents were ate least somewhat concerned about a company's use of emotional data from targeted advertisement, which supports the motivation for this work. Second, the EON played by a DARE-GP system was either noninvasive or completely inaudible for nearly all respondents.  Moreover, none of the respondents stated that they would be unwilling to DARE-GP. Finally, the record/playback solution, which is an analog for how any of the non-real-time systems would need to be deployed, was significantly less appealing; nearly half of the respondents stated that they would be unwilling to use this system. 

\begin{table}
\caption{Survey response counts for all respondents}
\begin{tabular}{|l|c|c|c|c|c|}
\hline
\textbf{Question 1}&Very Unconcerned&Somewhat Unconcerned&Neutral&Somewhat Concerned&Very Concerned \\
& $0$ & $1$ & $1$ & $8$ & $9$ \\
\hline
\textbf{Question 2}&Very Distracting&Somewhat Distracting&Neutral&Noninvasive&Completely Inaudible \\
& $0$ & $1$ & $1$ & $8$ & $9$ \\
\hline
\textbf{Question 3}&Very Unwilling&Somewhat Unwilling&Neutral&Somewhat Willing&Very Willing \\
& $0$ & $0$ & $0$ & $9$ & $10$ \\
\hline
\textbf{Question 4}&Very Unwilling&Somewhat Unwilling&Neutral&Somewhat Willing&Very Willing \\
& $2$ & $7$ & $3$ & $4$ & $3$ \\
\hline
\end{tabular}
\label{table:surveyresponses}
\end{table}

\end{document}